\pdfoutput=1

\documentclass[11pt]{article}

\usepackage[]{acl}

\usepackage{times}
\usepackage{latexsym}

\usepackage[T1]{fontenc}

\usepackage[utf8]{inputenc}

\usepackage{booktabs}
\usepackage{float}
\newcommand{\cdb}{{CARE$_{db}$}}
\newcommand{\carego}{{CARE$^{G}$}}
\newcommand{\cbert}{{\sc care-bert}}
\usepackage[linesnumbered,ruled,vlined]{algorithm2e}
\usepackage{graphicx} 

\usepackage{microtype}

\title{``That's so cute!'': The CARE Dataset for Affective Response Detection}

\author{Jane Dwivedi-Yu \\
  Meta AI\\
  \texttt{janeyu@fb.com} \\\And
  Alon Y. Halevy \\
  Meta AI\\
  \texttt{ayh@fb.com} \\}

\begin{document}

\maketitle

\begin{abstract}
Social media plays an increasing role in our communication with friends and family, and our consumption of information and entertainment. Hence, to design effective ranking functions for posts on social media, it would be useful to predict the affective response to a post (e.g., whether the user is likely to be humored, inspired, angered, informed). Similar to work on emotion recognition (which focuses on the affect of the publisher of the post), the traditional approach to recognizing affective response would involve an expensive investment in human annotation of training data. 

We create and publicly release \cdb, a dataset of 230k social media annotations according to 7 affective responses using the Common Affective Response Expression (CARE) method. The CARE method is a means of leveraging the signal that is present in comments that are posted in response to a post, providing high-precision evidence about the affective response to the post without human annotation. Unlike human annotation, the annotation process we describe here can be iterated upon to expand the coverage of the method, particularly for new affective responses. We present experiments that demonstrate that the CARE annotations compare favorably with crowdsourced annotations. Finally, we use \cdb\ to train  competitive BERT-based models for predicting affective response as well as emotion detection, demonstrating the utility of the dataset for related tasks. 

\end{abstract}

\section{Introduction}

Social media and other online media platforms have become a common means of both interacting and connecting with others as well as finding interesting, informing, and entertaining content. Users of those platforms depend on the ranking systems of the recommendation systems to show them information they will be most interested in and to safeguard them against unfavorable experiences. 

\begin{figure}[H]
  \centering
  \includegraphics[width=0.95\linewidth]{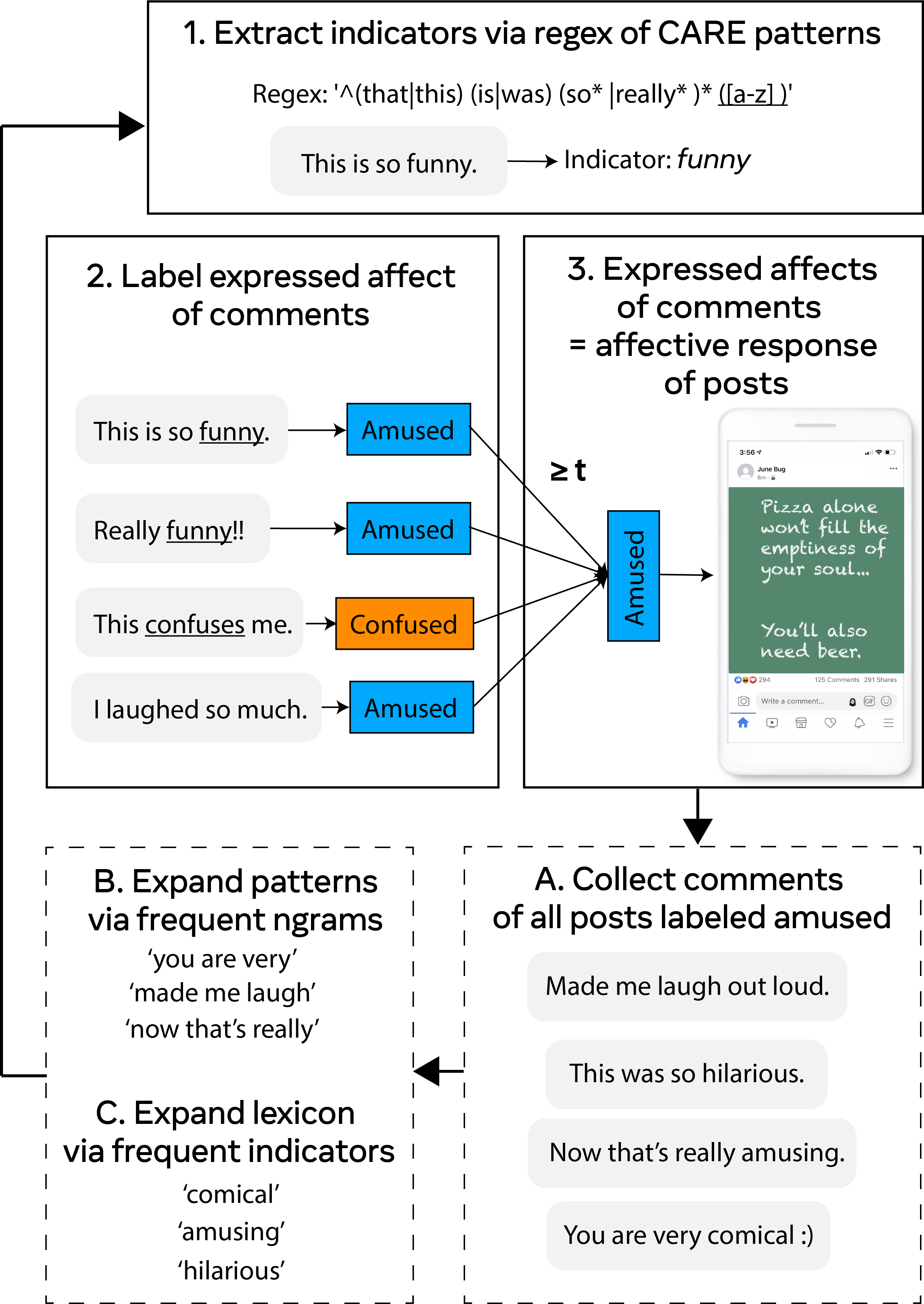} 
  \caption{Overview of the CARE Method (pseudo-code in Appendix, Algorithm~\ref{alg:expand}). The top half of the figure (steps 1--3) shows how the affective response to a post is computed by aggregating the expressed affects in comments from users viewing the post. The bottom half of the figure (steps A--C) shows how we expand the collection of CARE patterns and the lexicon based on labels that have been obtained from prior iterations. }
  
  \label{fig:care_diagram}
\end{figure}

Towards this end, a key technical problem is
to predict the \textit{affective response} that a user may have when they see a post. Some affective responses can be described by emotions (e.g., angry, joyful), and others may be described more as  experiences (e.g., entertained, inspired). 
Predicting affective response differs from emotion detection in that the latter focuses on the emotions expressed by the publisher of the post (referred to as the \textit{publisher affect} in~\citet{chen2014predicting}) and not on the viewer of the content. While the publisher's emotion may be relevant to the affective response, it only provides a partial signal \cite{dwivediyu2022}, and the two are not always equivalent (see Figure~\ref{fig:clown} for an illustrative example). Affective response for recommender systems has shown to be critical in several applications such as music, emotional health monitoring systems, product and travel recommendations \cite{rosa2015music, rosa2018knowledge, akram2020choseamobile, artemenko2020using, dwivediyu2022}.

Current approaches to predicting affective response require obtaining training data from human annotators who try to classify content into classes of a given taxonomy. However, obtaining enough training data can be expensive, and moreover, due to the subjective nature of the problem, achieving consensus among annotators can be challenging. Some methods explore inferring responses from physiological data or facial expressions from users, but this is a highly invasive process and can be difficult to scale to multiple users. \cite{tkalvcivc2017research, tkalvcivc2019prediction, angelastro2019predicting}.

This paper introduces the Common Affective Response Expression method (CARE for short), a means of obtaining labels for affective response in an unsupervised way from the comments written in response to online posts. CARE uses patterns and a keyword-affect mapping to identify expressions in comments that provide high-precision evidence about the affective response of the readers to the post. For example, the expression ``What a hilarious story'' may indicate that a post is humorous and ``This is so cute'' may indicate that a post is adorable.  We seed the system with a small number of high-precision patterns and mappings. We then iteratively expand on the initial set by considering frequent patterns and keywords in unlabeled comments on posts labeled by the previous iteration.

Using CARE, we create the largest dataset to date for affective response, \cdb, which contains 230k posts annotated according to 7 affective responses. We validate the effectiveness of CARE by comparing the CARE annotations with crowd-sourced annotations. Our experiments show that there is a high degree of agreement between the annotators and the labels proposed by CARE (e.g., in 90\% of the cases, at least two out of three annotators agree with all the CARE labels). Furthermore, we show that the CARE patterns/lexicon have greater accuracy than applying SOTA emotion recognition techniques to the comments. Using \cdb, we train \cbert\footnote{The CARE patterns, lexicon, \cdb, and \cbert\ are made available on github.}, a BERT-based model that can predict affective response {\em without} relying on comments. \cbert\ provides strong baseline performance for the task of predicting affective response, on par with the SOTA models for emotion recognition. Furthermore, we show that \cbert\ can be used for transfer learning to a different emotion-recognition task, achieving similar performance to~\citet{demszky2020goemotions}, which relied on manually-labeled training data. 


 

\section{Related work}
\label{section:related}

We first situate our work with respect to previous research on related tasks.

\subsection{Emotion detection in text}
Approaches to emotion detection can be broadly categorized into three groups: lexicon-based, machine learning, and combinations of the first two. The lexicon-based approach typically leverages lexical resources such as lexicons and encoded rules to guide emotion prediction \citep{tao2004context, ma2005emotion, asghar2017sentence}. Though these methods can be fast and interpretable, they are often not as robust and flexible because of the constraints of the lexicon \citep{alswaidan2020survey, acheampong2020text}. Additionally, the scope of emotions predicted by these works is usually fairly small, ranging from two to five, and most datasets utilized are usually smaller than 10k, making it unclear if they extrapolate well. Among the ML approaches, many SOTA works employ deep learning methods \citep{demszky2020goemotions, felbo2017using,barbieri2018semeval,huang2019ana,baziotis2017datastories,huang2019emotionx}, but while these show significant improvement over prior techniques, they are highly uninterpretable and often require prohibitively large human-labeled datasets to train. In both the lexicon-based approach and the ML-approach, the classes of emotions predicted in these works are usually non-extendable or require additional labeled data. 

While there are some commonalities between works in emotion detection and affective response detection, the problems are distinct enough that we cannot simply apply emotion recognition techniques to our setting. Emotion recognition focuses on the publisher affect (the affect of the person writing the text). The publisher affect may provide a signal about the affective response of the reader, but there is no simple mapping from one to the other. For example, being `angered' is an affective response that does not only result from reading an angry post---it can result from a multitude of different publisher affects (e.g. excited, angry, sympathetic, embarrassed, or arrogant). For some affective responses, such as being `grateful' or feeling `connected' to a community, the corresponding publisher affect is highly unclear.

\subsection{Affective response detection}
There have been some works that address affective response in limited settings, such as understanding reader responses to online news~\citep{katz2007swat, strapparava2007semeval, lin2008emotion, lei2014towards}. In contrast, our goal is to address the breadth of content on social media. There are works which use Facebook reactions as a proxy for affective response, but these are constrained by the pre-defined set of reactions ~\citep{clos2017predicting, raad2018aseds, pool2016distant, graziani2019jointly, krebs2017social}. The work described in~\citet{rao2014affective} and~\citet{bao2011mining} attempts to associate emotions with {\em topics}, but a single topic can have a large variety of affective responses when seen on social media, and therefore their model does not apply to our case. Some works in the computer vision community study affective response to images~\citep{chen2014predicting, jou2014predicting}; as they note, most of the work in the vision community also focuses on  publisher affect. 
 
\subsection{Methods for unsupervised labeling}
\label{sec:weak_labeling}
A major bottleneck in developing models for emotion and affective response detection is the need for large amounts of training data. As an alternative to manually-labeled data, many works utilize metadata such as hashtags, emoticons, and Facebook reactions as pseudo-labels~\citep{wang2012harnessing, suttles2013distant, hasan2014using, mohammad2015using}. However, these can be highly noisy and limited in scope. For example, there exist only seven Facebook reactions, and they do not necessarily correspond to distinct affective responses. Additionally, for abstract concepts like emotions, hashtagged content may only capture a superficial interpretation of the concept. For example, \#inspiring on Instagram will give many photos featuring selfies or obvious inspirational quotes, which do not sufficiently represent inspiration. The work we present here extracts labels from free-form text in comments rather than metadata. The work done in~\citet{sintsova2016dystemo} is similar to our work in that it pseudo-labels tweets and extends its lexicon, but the classifier itself is a keyword, rule-based approach and is heavily reliant on the capacity of these lexicons. In contrast, our work leverages the high precision of CARE and uses these results to train a model, which is not constrained by the lexicon size in its predictions. Our method also employs bootstrapping to expand the set of patterns and lexicon, similar to \citet{agichtein2000snowball} and \citet{jones1999bootstrapping} but focuses on extracting affect rather than relation tuples.

\section{The CARE Method}
\label{sec:length}

In this section, we provide a formal description of CARE for annotating the affective response of posts. Before we proceed, we note two aspects of affective responses. First, there is no formal definition for what qualifies as an affective response. In practice, we use affective responses to understand the experience that the user has when seeing a piece of content, and these responses may be both emotional and cognitive. Second, the response a user may have to a particular piece of content is clearly a very personal one. Our goal here is to predict whether a piece of content is generally likely to elicit a particular affective response. In practice, if the recommendation system has models of user interests and behavior, these would need to be combined with the affect predictions. 


\subsection{CARE patterns and the CARE lexicon}
\label{sec:care-patterns}

CARE is composed of two major components: CARE patterns, regular expressions used to extract information from the comments of a post, and the CARE lexicon, a keyword-affect dictionary used to map the comment to an affect. 

CARE patterns are not class or affect-specific and leverage common structure present in comments for affective response extraction. There is an unlimited number of possible CARE patterns, but we seeded the system with six CARE patterns and an additional 17 more were automatically discovered using the expansion method.  In the same spirit as Hearst Patterns~\citep{hearst1992automatic}, CARE patterns are tailored to extract specific relationships and rely on two sets of sub-patterns:
\begin{itemize}
    \item Exaggerators $\{E\}$: words that intensify or exaggerate a statement, e.g.,  \textit{so}, \textit{very}, or \textit{really}. 
    \item Indicators $\{I\}$: words (up to 3) that exist in the CARE lexicon, which maps the indicator to a particular class. For example, `funny' in ``This is so funny'' would map to \textit{amused}. 
\end{itemize}

Consequently, one example of the six CARE patterns that were used to seed the system is the following: \{this$\vert$that$\vert$those$\vert$these\}\{is$\vert$are\}*\{$E$\}$^*$\{$I$\}$^+$\\ (the symbol $^*$ (resp.~$^+$) indicates that zero (resp.\ one) or more matches are required). An example of an instantiation of this pattern would be, `This is so amazing!'.

Given the indicators extracted by the CARE patterns, the CARE lexicon is responsible for mapping the comment to particular affective responses. The lexicon contains 163 indicators for the 7 classes we consider (123 of which were automatically identified in the expansion process described in the next section). We also considered using other lexicons~\citep{strapparava2004wordnet,poria2014emosenticspace,staiano2014depechemood,esuli2006sentiwordnet,mohammad2013nrc}, but we found that they were lacking enough application context to be useful in our setting. Table~\ref{tab:care_definitions} shows the affects in the CARE lexicon and corresponding definitions and example comments that would fall under each affect (or class). The classes \textit{excited}, \textit{angered}, \textit{saddened}, and \textit{scared} were chosen since they are often proposed as the four basic emotions \cite{wang2011drosophila, jack2014dynamic, gu2016neuromodulator, zheng2016safety}. The classes \textit{adoring}, \textit{amused}, and \textit{approving} were established because they are particularly important in the context of social media for identifying positive content that users enjoy.  Overall, a qualitative inspection indicated that these seven have minimal conceptual overlap and sufficiently broad coverage. We note, however, that one of the benefits of the method we describe is that it is relatively easy to build a model for a new class of interest compared to the process of human annotation.


\begin{table*}[htb]
    \centering
    \small
    \begin{tabular}{lllc}
        \toprule
        \textbf{AR} & \textbf{Definition} & \textbf{Example comment} & \textbf{Size}\\
        \midrule
        Adoring & Finding someone or something cute, adorable, or attractive. & \textit{He is the cutest} thing ever. & 36\\
        Amused & Finding something funny, entertaining, or interesting. & \textit{That was soooo funny.} & 30\\
        Approving & Expressing support, praise, admiration, or pride. &  \textit{This is really fantastic!} & 102\\
        Excited & Expressing joy, zeal, eagerness, or looking forward to something. & \textit{Really looking forward to} this! & 41\\
        Angered & Expressing anger, revulsion, or annoyance. & \textit{I'm so frustrated} to see this. & 26\\
        Saddened & Expressing sadness, sympathy, or disappointment. & \textit{So sad from reading this}. & 34\\ 
        Scared & Expressing worry, concern, stress, anxiety, or fear. & \textit{Extremely worried} about finals. & 2\\
        \bottomrule
    \end{tabular}
    \caption{Definition of affective responses (AR), examples of comments which would map to each affective response, and the number of posts (in thousands) per class in \cdb. The portion of each example which would match a CARE pattern in a reg-ex search is italicized.}
    \label{tab:care_definitions}
\end{table*}

\subsection{Labeling posts}
\label{sec:post-labeling}

Here we describe how to combine and use the two major components (CARE patterns and lexicon) at the comment-level in order to annotate the post-level affective response. The pipeline for labeling posts is shown in steps 1--3 of Figure~\ref{fig:care_diagram} and described in detail in Algorithm~\ref{alg:expand} in the Appendix. We begin with reg-ex matching of CARE patterns and individual sentences of the comments.  We truncate the front half of a sentence if it contains words like `but' or `however' because the latter half usually indicates their predominant sentiment. We also reject indicators that contain negation words such as `never', `not', or `cannot' (although one could theoretically map this to the opposite affective response using Plutchik's Wheel of Emotions \citep{plutchik1980general}). Note that contrary to traditional rule-based or machine-learning methods, we do not strip stop words like `this' and `very' because it is often crucial to the regular expression matching, and this specificity has a direct impact on the precision of the pipeline. 

Given the reg-ex matches, we use the lexicon to map the indicators to the publisher affect of the comment (e.g., \textit{excited}). Because the expressed affect of the comments equates to the affective response of a post, we obtain a post-level affective response label by aggregating the comment-level labels and filtering out labels that have a support smaller than $t$. Specifically, a post would be labeled with the affective response $a$ if at least $t$ of the comments were labeled with $a$. In our experiments, we used a value of $t=5$, after qualitative inspection of \cdb, discussed in Section \ref{sec:evaluation}.

\paragraph{Expanding CARE patterns/lexicon:}
\label{sec:expansion}
We seeded our patterns and lexicon with a small intuitive set and then expanded them by looking at common n-grams that appear across posts with the same label (steps A--C of Figure~\ref{fig:care_diagram}). At a high level, for a given affect $a$, consider the set, $comm(a)$, of all the comments on posts that were labeled $a$, but did not match any CARE pattern.  From these comments, we extract new keywords (e.g. `dope' for \textit{approving} as in `This is so dope.') for the CARE lexicon by taking the most frequent n-grams in $comm(a)$ but infrequently in $comm(b)$, where b includes all classes except a. On the other hand, the most common n-grams co-occuring with multiple classes were converted to regular expressions and then added as new CARE patterns (see Table~\ref{tab:ngram_examples} for a few examples). We added CARE patterns according to their frequency and stopped when we had sufficient data to train our models. After two expansion rounds, the set of patterns and indicators increased from 6 to 23 and 40 to 163, respectively. Counting the possible combinations of patterns and indicators, there are roughly 3500 distinct expressions. \textit{When considering the possible 23 CARE patterns, 163 CARE lexicon indicators, and 37 exaggerators, there are a total of 130k possible instantiations of a matching comment.}

\section{Evaluating \cdb}
\label{sec:evaluation}
 
In this section we apply our method to social media posts and validate these annotations using human evaluation (Section~\ref{sec:annotation}). Section~\ref{sec:error_analysis} discusses class-wise error analysis, and in Section~\ref{sec:carego}, we explore the alternative possibility of creating \cdb\ using a SOTA publisher-affect classifier \cite{demszky2020goemotions} to label the comments.
 
 \paragraph{\cdb:} 
 
 Our experiments use a dataset that is created from Reddit posts and comments in the pushshift.io database that were created between 2011 and 2019. We create our dataset, \cdb, as follows. We used CARE patterns and the CARE lexicon to annotate 34 million comments from 24 million distinct posts. After filtering with a threshold of $t=5$, we obtained annotations for 400k posts (the total number of posts that have at least 5 comments was 150 million). The low recall is expected given the specificity of CARE patterns/lexicon. We also filtered out posts that have less than 10 characters, resulting in a total of 230k posts in \cdb. Table~\ref{tab:care_definitions} shows the breakdown of cardinality per affective response. 195k of the posts were assigned a single label, whereas 26k (resp.\ 8k) were assigned two (resp.\ three) labels. Note that the distribution of examples per class in \cdb\  is not reflective of the distribution in the original data, because different classes have different recall rates. The \cdb\ dataset features the pushshift.io id and text of the post as well as the annotations using CARE.

\subsection{Human evaluation}
\label{sec:annotation}

In our next experiment, we evaluate the labels predicted by CARE with the help of human annotators using Amazon Mechanical Turk (AMT), restricting to those who qualify as AMT Masters and having lifetime approval rating greater than 80\%. The dataset for annotation was created as follows. We sub-sampled a set of 6000 posts from \cdb, ensuring that we have at least 700 samples from each class and asked annotators to label the affective response of each post. Annotators were encouraged to select as many as appropriate and also permitted to choose `None of the above' as shown in Figure~\ref{fig:amt_ui}. In addition to the post, we also showed annotators up to 10 sampled comments from the post in order to provide more context. Every post was shown to three of the 91 distinct annotators. For quality control, we also verified that there no individual annotator provided answers that disagreed with the CARE labels more than 50\% of the time on more than 100 posts. 

We observed an average Fleiss' kappa score 0.59, which is considered moderate to high agreement, the breakdown of which is shown in Table~\ref{fig:amt_ui}. Table~\ref{tab:amt_results} shows that the rate of agreement between the annotators and the labels proposed by the CARE method is high. For example, 94\% of posts had at least one label proposed by CARE that was confirmed by 2 or more annotators, and 90\% had {\em all} the labels confirmed. The last column measures the agreement among annotators on labels that were not suggested by CARE, which was 53\% when confirmed by 2 or more annotators. We expected this value to be reasonably large because the CARE patterns/lexicon were designed to generate a highly precise set of labels, rather than highly comprehensive ones. However, the value is still much smaller relative to the agreement rate for the CARE labels. On average, each annotation answer contained around 1.8 labels per post (with a standard deviation of 0.9). We note that  `None of the above' was chosen less than 0.2\% of the time. Table~\ref{tab:amt_class_results} and Figure~\ref{fig:class_prevalence} present annotator agreement statistics and label prevalence, respectively, broken down by class. Figure~\ref{fig:corr_heatmap} shows the Spearman correlation between each class and a hierarchical clustering.

\begin{table}[htb]
\small
\centering
\begin{tabular}{cccc} 
\toprule
 \textbf{\# Agree} & \textbf{Any CARE} & \textbf{All CARE} & \textbf{Other} \\
\midrule
  $\ge 1$ & 98 & 96 & 82\\
  $\ge 2$ & 94 & 90 & 53\\
  = 3 & 80 & 76 & 24\\
  \bottomrule
  \hline
\end{tabular}
\caption{The rate of agreement between the annotators and the labels proposed by CARE.  The first column specifies the number of annotators to be used for consensus. The rest of the columns shows for all posts, the average rate of intersection of the human labels with at least one CARE label, all CARE labels, and any label that is not a CARE label.}
\label{tab:amt_results}
\end{table}

\subsection{Error Analysis}
\label{sec:error_analysis}

Evaluating CARE in the previous sections (Figure~\ref{tab:amt_class_results}, Table~\ref{tab:goemotions_class_results}) revealed that the accuracy of CARE varies by class and in particular, is lower for  \textit{amused} and \textit{excited}.
To better understand if certain pattern or indicator matches are at fault here, we investigate the precision and recall at the pattern and lexicon level. 

Recall that instantiating a match for a comment involves choosing a (pattern, keyword) combination. Separating the lexicon from the patterns enables us to encode a large number of instantiated patterns parsimoniously, but some pair combinations provide a much weaker signal than others, particularly for the class \textit{amused} (see Figure~\ref{fig:match_analysis}). Hence, for future iterations of CARE, we have implemented a mechanism to exclude certain pattern and keyword combinations and a means for using different thresholds for each class. 

\begin{figure}[t]
  \centering
  \includegraphics[width=0.87\linewidth]{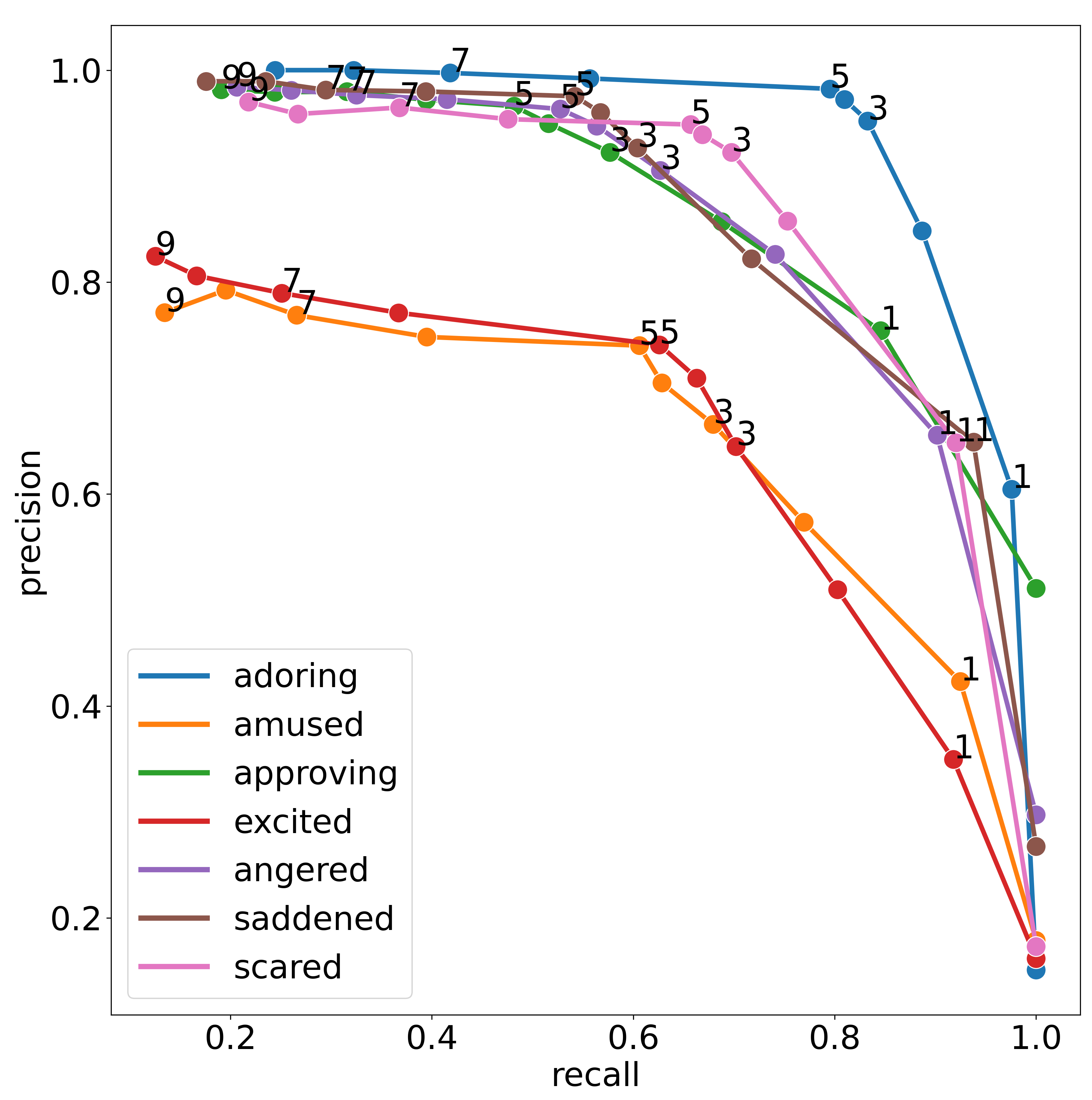} 
  \caption{Precision versus recall of each class using varying thresholds (t = 0 to 9). Ground truth labels utilized are those which have at least 2 out of 3 annotator agreement. For clarity, only odd values of t are labeled.}
  \label{fig:threshold_plot.png}
\end{figure}

Alternatively, another mechanism for accomodating these class-wise discrepancies in performance is by tuning for each class an optimal threshold $t$ (i.e., the number of matched comments we need to see in order to reliably predict a label). Figure~\ref{fig:threshold_plot.png} shows how the precision and recall of each class varies according to different threshold values.
To achieve precision and recall greater than 0.7, a threshold of 1 actually seems viable for most classes, while for \textit{amused} and \textit{excited} a threshold of at least 3 is needed. In fact, for most of the classes, using thresholds larger than 3 has negligible impact on the precision score, but does reduce the recall.

\subsection{Can we leverage emotion classification?}
\label{sec:carego}

Recall, steps 1 and 2 of Figure~\ref{fig:care_diagram} uses the CARE patterns and lexicon to label the publisher affect of the comments. Conceivably, this could have been done instead by using a SOTA emotion classifier such as the GoEmotions classifier \cite{demszky2020goemotions}, which is trained specifically to predict the publisher affect of Reddit comments. Here we show that our method for labeling the publisher affect of comments performs comparatively better. Let us define the method \carego, a modified version of the CARE method where steps 1 and 2 are replaced with labels using the GoEmotions classifer. We apply \carego to our human annotated dataset ~(Section~\ref{sec:annotation}) by first applying the GoEmotions classifier to all comments of the posts. These GoEmotion labels are then mapped to our taxonomy in Table~\ref{tab:care_definitions} using the mapping defined in Table~\ref{tab:goemotions_class_results}, which is based on the grouping of emotions at the Ekman level used in \citet{demszky2020goemotions}.  We then, as usual, aggregate and filter post labels according to a threshold $t$. 

\carego\ (Table~\ref{tab:amt_results_compare}) shows a relative decrease of 12.9\% and 18.0\% in the rate of annotator agreement with any and all labels, respectively, compared to that of CARE. These decreases hold even when partitioning on each individual class. The comparatively lower performance of \carego~is most likely due to the low F1-scores ($<$0.4) of the GoEmotions classifer for nearly half of the 28 classes, as reported in the original work~\citet[Table~4]{demszky2020goemotions}. 
It is also important to note that in addition to demonstrating higher precision, CARE patterns and lexicon are valuable because they do not require human annotated data, unlike GoEmotions. It may, however, be useful to leverage multiple emotion detection approaches and Section~\ref{sec:care_g} discusses a potential ensembling strategy for this.

To validate the mapping in Table~\ref{tab:goemotions_class_results}, we applied steps 1 and 2 of CARE to the GoEmotions dataset (see Section~\ref{sec:goemotions}), and computed the rate of agreement among the labels in our defined mapping. We find this rate of agreement to be high (87.3\% overall). Note, we perform this equivalence at the publisher affect level, because as discussed before, the affective response and publisher affect are not always equivalent. In addition to prior work \cite{dwivediyu2022}, Section~\ref{sec:response_and_publisher} presents experiments that indicate that affective response and publisher affect labels intersect only 44\% of the time.

\begin{table}[h]
\centering
\small
\begin{tabular}{ccc} 
\toprule
 \textbf{AR} & \textbf{GoEmotion label} & \textbf{\% agree} \\
\midrule
  Amused & Amusement & 79.8 \\
  Approving & Admiration, approval & 89.3 \\
  Excited & Joy & 81.3 \\
  Angered & Anger, Annoyance, Disgust & 93.3 \\
  Saddened & Disappointment, sadness & 90.9\\
  Scared & Fear, nervousness & 84.9\\
  \bottomrule
  \hline
\end{tabular}
\caption{CARE to GoEmotions mapping. The last column summarizes the rate at which the mapping holds. The average across all datapoints was 87.3\%.}
\label{tab:goemotions_class_results}
\end{table}

\section{Predicting affective response for posts without comments}
\label{sec:modeling}

In this section we describe \cbert, a multi-label affective response classifier that is trained only on the post-level text and annotations in \cdb. Such a model is important in order to make predictions early in the life of the post and in cases where the comments may not match any CARE patterns or keywords. Note that \textit{the model is not given the comments} text and is therefore not restricted to the CARE pattern semantics.  In section~\ref{sec:tranfer_learning}, we describe how \cbert\ can be further fine-tuned for related tasks like emotion detection.

\subsection{Creating and evaluating CARE-BERT}



We train \cbert\ with the CARE labels in \cdb, using the pre-trained model bert-base-uncased in the Huggingface library~\citep{wolf2019huggingface}. We use a max length of 512 and we add a dropout layer with a rate of 0.3 and a dense layer to allow for multi-label classification. We used an Adam optimizer with a learning rate of 5e-5, a batch size of 16, and 5 epochs. We used a train/validation/test split of 80/10/10\%. See Section~\ref{sec:bert_setup_details} for other settings we explored.

The evaluation on the human-annotated set (held out from training) is shown in Table~\ref{tab:bert_results_annotated}. We use labels with support from all annotators as ground truth. The classes of lowest prevalence, such as \textit{scared}, had the poorest results, while the more frequent classes (\textit{adoring}, \textit{approving}, \textit{saddened}) had the highest results. To put these results in perspective, we use the mapping in Table~\ref{tab:goemotions_class_results} and compare with the numbers from \citet{demszky2020goemotions}. Note, the comparison is \textit{not} for the same dataset---our results pertain to predicting on the post whereas GoEmotions predicts the comments. Still, \cbert~demonstrates a 35\% improvement in the overall micro-averaged F1-score.

\paragraph{CARE vs. CARE-BERT:} Compared to the human annotators and CARE, \cbert\ is disadvantaged by not having access to the comments. We use human annotated set of \cdb~ and find that 0.89 of the CARE labels are also proposed by human annotators, while this value is 0.72 for \cbert~(Table~\ref{tab:care_bert_stats}). In Table~\ref{tab:care_bert_examples} we display select examples that may illustrate reasons for this discrepancy. Firstly, one of the challenges that \cbert\  faces is that there may not be sufficient context in the post alone. In the example ``Who is this LIRIK guy, and why does he have 50K subscribers'' it is challenging to predict that some people find the subject adorable without additional context. Relatedly, the conversation that the post initiates can be challenging to foresee. The last example reads "AskReddit: Imagine the last thing you ate has been made illegal. What would that be?" In some cases, commenters ate something they didn't like and are therefore content with the premise. In other cases, commenters ate something they very much enjoy and are saddened by the hypothetical. Our results show that this is not particular to `AskReddit' posts, and given these challenges, it is reasonable that the CARE method provides more reliable labels.




\begin{table}[thb]
\centering
\small
\begin{tabular}{ccccc} 
\toprule
\textbf{AR} &\textbf{P} & \textbf{R} & \textbf{F1} & \textbf{GoEmotions F1}\\
\midrule
  Adoring & 0.73 & 0.66 & 0.70 & -\\
  Amused & 0.63 & 0.54 & 0.60 & \textbf{0.80}\\
  Approving & 0.73 & 0.72 & \textbf{0.75} & 0.53\\
  Excited & 0.58 & 0.52 & \textbf{0.58} & 0.51\\
  Angered & 0.70 & 0.61 & \textbf{0.69} & 0.40\\
  Saddened & 0.78 & 0.62 & \textbf{0.73} & 0.39\\
  Scared & 0.68 & 0.3 & 0.47 & \textbf{0.54}\\
  \midrule
  micro-avg & 0.70 & 0.68 & \textbf{0.69} & 0.51\\
  macro-avg & 0.69 & 0.62 & \textbf{0.65} & 0.53\\
  stdev & 0.06 & 0.12 & 0.09 & 0.14\\
  \bottomrule
\end{tabular}
\caption{Precision (P), recall (R), and F1 of \cbert~using \cdb~on the post text of the human-annotated set and F1-scores of the GoEmotions classifier from \citet{demszky2020goemotions} on comments.}
\label{tab:bert_results_annotated}
\end{table}

\subsection{Transfer learning to emotion detection}
\label{sec:tranfer_learning}

We now demonstrate that \cbert\ is  also useful for pre-training of another related task in a setting with limited annotated data. We consider transfer learning to the ISEAR Dataset~\citep{scherer1994evidence}, which is a collection of  7666 statements from a diverse set of 3000 individuals labeled according to six categories (anger, disgust, fear, guilt, joy, sadness, and shame). The labels pertain to the \textit{publisher affect} and not affective response, as considered in this work. Our experiment explores transfer learning to predict the labels in the ISEAR dataset using an additional drop-out layer of 0.3 and a dense layer. 

\begin{figure}[t]
  \centering
  \includegraphics[width=0.83\linewidth]{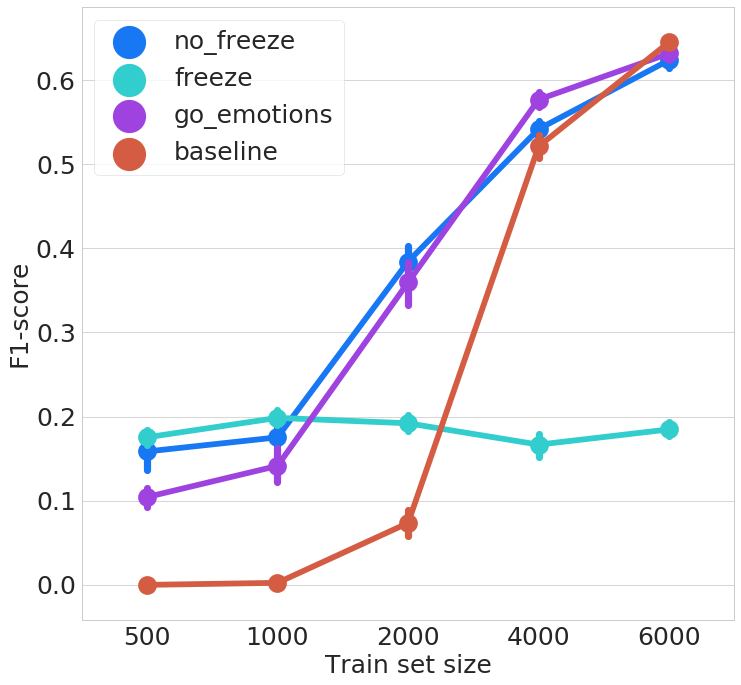} 
 
  \caption{The F1-score of each model using varying training set sizes of the ISEAR dataset. The light blue line refers to using \cbert, but with freezing all parameters except in the last layer. The dark blue is the same but without freezing. Lastly, the purple line refers to the same architecture as \cbert~(no freezing) but trained on GoEmotions instead of \cdb, and the red line is trained only on the ISEAR dataset using a bert-base-uncased model with the same hyperparameters.}
  \label{fig:transfer_learning}
\end{figure}

Our experiments follow closely to that of \citet{demszky2020goemotions} and uses different training set sizes (500, 1000, 2000, 4000, and 6000) for 10 different train-test splits. We plot the average and standard deviation in the F1-scores across these 10 splits in Figure~\ref{fig:transfer_learning}. We compare four different fine-tuning setups: the first two are trained using \cbert~and then fine-tuned on the benchmark dataset, one with no parameter freezing (no\_freeze), and one with all layers but the last two frozen (freeze). The third setup is similar to \cbert\ (no\_freeze) but is trained on GoEmotions rather than \cdb. The last setup is the bert-base-uncased model trained only on ISEAR, where all setups use the same architecture and hyperparameters as discussed in Section~\ref{sec:modeling}. 

Our values differ slightly from that cited in \citet{demszky2020goemotions} due to the small differences in architecture and hyperparameters. However, the overall results corroborate that of \citet{demszky2020goemotions} in that models with additional pre-training perform better than the baseline (no additional pre-training) for limited sample sizes. From Figure~\ref{fig:transfer_learning}, it is apparent that \cbert\ and the model built from GoEmotions perform essentially on par in these transfer learning experiments in spite of the fact that \cbert\ does not utilize human annotations. It is also worth noting that GoEmotions and the ISEAR dataset address the same task (emotion detection) while \cbert\ predicts affective response. The comparable performance of \cbert\ with the GoEmotions models demonstrates the utility of \cbert\ for other tasks with limited data and the promise of CARE as a means of reliable unsupervised labeling.

\section{Conclusion}

We described a method for extracting training data for models that predicts the affective response to a post on social media. CARE is an efficient, accurate, and scalable way of collecting unsupervised labels and can be extended to new classes. Using CARE, we created \cdb, a large dataset which can be used for affective response detection and other related tasks, as demonstrated by the competitive performance of \cbert\ to similar BERT-based models in emotion detection. We release the dataset and models in the hopes that this will unlock future research.

In particular, there are two main cases in which CARE can be improved upon: (1) when there does not exist a set of common phrases that are indicative of an affect, and (2) when an indicator maps to multiple affects. In the latter case, there is still partial information that can be gleaned from the labels. In addition to developing methods for the above cases, future work also includes incorporating emojis, negations, and punctuation, and extending to new classes.  Finally, we also plan to investigate the use of CARE for predicting the affective response to images as well as multi-modal content such as memes.




\bibliography{anthology,care}

\begin{thebibliography}{56}
\expandafter\ifx\csname natexlab\endcsname\relax\def\natexlab#1{#1}\fi

\bibitem[{dwi()}]{dwivediyu2022}


\bibitem[{Acheampong et~al.(2020)Acheampong, Wenyu, and
  Nunoo-Mensah}]{acheampong2020text}
Francisca~Adoma Acheampong, Chen Wenyu, and Henry Nunoo-Mensah. 2020.
\newblock \href {https://doi.org/10.1002/eng2.12189} {Text-based emotion
  detection: Advances, challenges, and opportunities}.
\newblock \emph{Engineering Reports}, page e12189.

\bibitem[{Agichtein and Gravano(2000)}]{agichtein2000snowball}
Eugene Agichtein and Luis Gravano. 2000.
\newblock \href {https://doi.org/10.1145/336597.336644} {Snowball: Extracting
  relations from large plain-text collections}.
\newblock In \emph{Proceedings of the fifth ACM conference on Digital
  libraries}, pages 85--94.

\bibitem[{Akram et~al.(2020)Akram, Hussain, Toure, Yang, and
  Jalal}]{akram2020choseamobile}
Sheeraz Akram, Shariq Hussain, Ibrahima~Kalil Toure, Shunkun Yang, and Humza
  Jalal. 2020.
\newblock Choseamobile: A web-based recommendation system for mobile phone
  products.
\newblock \emph{Journal of Internet Technology}, 21(4):1003--1011.

\bibitem[{Alswaidan and Menai(2020)}]{alswaidan2020survey}
Nourah Alswaidan and Mohamed Menai. 2020.
\newblock \href {https://doi.org/10.1007/s10115-020-01449-0} {A survey of
  state-of-the-art approaches for emotion recognition in text}.
\newblock \emph{Knowledge and Information Systems}, 62.

\bibitem[{Angelastro et~al.(2019)Angelastro, Carolis, and
  Ferilli}]{angelastro2019predicting}
Sergio Angelastro, B~Carolis, and Stefano Ferilli. 2019.
\newblock Predicting user preference in pairwise comparisons based on emotions
  and gaze.
\newblock In \emph{International Conference on Industrial, Engineering and
  Other Applications of Applied Intelligent Systems}, pages 253--261. Springer.

\bibitem[{Artemenko et~al.(2020)Artemenko, Pasichnyk, Kunanets, and
  Shunevych}]{artemenko2020using}
Olga Artemenko, Volodymyr Pasichnyk, Nataliia Kunanets, and Khrystyna
  Shunevych. 2020.
\newblock Using sentiment text analysis of user reviews in social media for
  e-tourism mobile recommender systems.
\newblock In \emph{COLINS}, pages 259--271.

\bibitem[{Asghar et~al.(2017)Asghar, Khan, Bibi, Kundi, and
  Ahmad}]{asghar2017sentence}
Dr.~Muhammad Asghar, Aurangzeb Khan, Afsana Bibi, Fazal Kundi, and Hussain
  Ahmad. 2017.
\newblock \href {https://doi.org/10.1007/s12559-017-9503-3} {Sentence-level
  emotion detection framework using rule-based classification}.
\newblock \emph{Cognitive Computation}, 9:1--27.

\bibitem[{Bao et~al.(2012)Bao, Xu, Zhang, Yan, Su, Han, and Yu}]{bao2011mining}
Shenghua Bao, Shengliang Xu, Li~Zhang, Rong Yan, Zhong Su, Dingyi Han, and Yong
  Yu. 2012.
\newblock \href {https://doi.org/10.1109/TKDE.2011.188} {Mining social emotions
  from affective text}.
\newblock \emph{IEEE Transactions on Knowledge and Data Engineering},
  24(9):1658--1670.

\bibitem[{Barbieri et~al.(2018)Barbieri, Camacho-Collados, Ronzano,
  Espinosa-Anke, Ballesteros, Basile, Patti, and Saggion}]{barbieri2018semeval}
Francesco Barbieri, Jose Camacho-Collados, Francesco Ronzano, Luis
  Espinosa-Anke, Miguel Ballesteros, Valerio Basile, Viviana Patti, and Horacio
  Saggion. 2018.
\newblock \href {https://doi.org/10.18653/v1/S18-1003} {{S}em{E}val 2018 task
  2: Multilingual emoji prediction}.
\newblock In \emph{Proceedings of The 12th International Workshop on Semantic
  Evaluation}, pages 24--33, New Orleans, Louisiana. Association for
  Computational Linguistics.

\bibitem[{Baziotis et~al.(2017)Baziotis, Pelekis, and
  Doulkeridis}]{baziotis2017datastories}
Christos Baziotis, Nikos Pelekis, and Christos Doulkeridis. 2017.
\newblock \href {https://doi.org/10.18653/v1/S17-2126} {{D}ata{S}tories at
  {S}em{E}val-2017 task 4: Deep {LSTM} with attention for message-level and
  topic-based sentiment analysis}.
\newblock In \emph{Proceedings of the 11th International Workshop on Semantic
  Evaluation ({S}em{E}val-2017)}, pages 747--754, Vancouver, Canada.
  Association for Computational Linguistics.

\bibitem[{Chen et~al.(2014)Chen, Chen, Hsu, Liao, and
  Chang}]{chen2014predicting}
Yan-Ying Chen, Tao Chen, Winston~H Hsu, Hong-Yuan~Mark Liao, and Shih-Fu Chang.
  2014.
\newblock \href {https://doi.org/10.1145/2578726.2578756} {Predicting viewer
  affective comments based on image content in social media}.
\newblock In \emph{proceedings of international conference on multimedia
  retrieval}, pages 233--240.

\bibitem[{Clos et~al.(2017)Clos, Bandhakavi, Wiratunga, and
  Cabanac}]{clos2017predicting}
J{\'e}r{\'e}mie Clos, Anil Bandhakavi, Nirmalie Wiratunga, and Guillaume
  Cabanac. 2017.
\newblock \href {https://doi.org/10.1007/978-3-319-56608-5_44} {Predicting
  emotional reaction in social networks}.
\newblock In \emph{European Conference on Information Retrieval}, pages
  527--533. Springer.

\bibitem[{Cox and Cox(2008)}]{cox2008multidimensional}
Michael~AA Cox and Trevor~F Cox. 2008.
\newblock Multidimensional scaling.
\newblock In \emph{Handbook of data visualization}, pages 315--347. Springer.

\bibitem[{Demszky et~al.(2020)Demszky, Movshovitz-Attias, Ko, Cowen, Nemade,
  and Ravi}]{demszky2020goemotions}
Dorottya Demszky, Dana Movshovitz-Attias, Jeongwoo Ko, Alan Cowen, Gaurav
  Nemade, and Sujith Ravi. 2020.
\newblock \href {https://doi.org/10.18653/v1/2020.acl-main.372}
  {{G}o{E}motions: A dataset of fine-grained emotions}.
\newblock In \emph{Proceedings of the 58th Annual Meeting of the Association
  for Computational Linguistics}, pages 4040--4054, Online. Association for
  Computational Linguistics.

\bibitem[{Esuli and Sebastiani(2006)}]{esuli2006sentiwordnet}
Andrea Esuli and Fabrizio Sebastiani. 2006.
\newblock \href {http://www.lrec-conf.org/proceedings/lrec2006/pdf/384_pdf.pdf}
  {{SENTIWORDNET}: A publicly available lexical resource for opinion mining}.
\newblock In \emph{Proceedings of the Fifth International Conference on
  Language Resources and Evaluation ({LREC}{'}06)}, Genoa, Italy. European
  Language Resources Association (ELRA).

\bibitem[{Felbo et~al.(2017)Felbo, Mislove, S{\o}gaard, Rahwan, and
  Lehmann}]{felbo2017using}
Bjarke Felbo, Alan Mislove, Anders S{\o}gaard, Iyad Rahwan, and Sune Lehmann.
  2017.
\newblock \href {https://doi.org/10.18653/v1/D17-1169} {Using millions of emoji
  occurrences to learn any-domain representations for detecting sentiment,
  emotion and sarcasm}.
\newblock In \emph{Proceedings of the 2017 Conference on Empirical Methods in
  Natural Language Processing}, pages 1615--1625, Copenhagen, Denmark.
  Association for Computational Linguistics.

\bibitem[{Graziani et~al.(2019)Graziani, Melacci, and
  Gori}]{graziani2019jointly}
Lisa Graziani, Stefano Melacci, and Marco Gori. 2019.
\newblock \href {https://doi.org/10.1007/978-3-030-30490-4_16} {Jointly
  learning to detect emotions and predict facebook reactions}.
\newblock In \emph{International Conference on Artificial Neural Networks},
  pages 185--197. Springer.

\bibitem[{Gu et~al.(2016)Gu, Wang, Wang, and Huang}]{gu2016neuromodulator}
Simeng Gu, Wei Wang, Fushun Wang, and Jason~H Huang. 2016.
\newblock Neuromodulator and emotion biomarker for stress induced mental
  disorders.
\newblock \emph{Neural plasticity}, 2016.

\bibitem[{Hasan et~al.(2014)Hasan, Agu, and Rundensteiner}]{hasan2014using}
Maryam Hasan, Emmanuel Agu, and Elke Rundensteiner. 2014.
\newblock \href {http://web.cs.wpi.edu/~emmanuel/publications/PDFs/C25.pdf}
  {Using hashtags as labels for supervised learning of emotions in twitter
  messages}.
\newblock In \emph{ACM SIGKDD workshop on health informatics, New York, USA}.

\bibitem[{Hearst(1992)}]{hearst1992automatic}
Marti~A. Hearst. 1992.
\newblock \href {https://www.aclweb.org/anthology/C92-2082} {Automatic
  acquisition of hyponyms from large text corpora}.
\newblock In \emph{{COLING} 1992 Volume 2: The 15th {I}nternational
  {C}onference on {C}omputational {L}inguistics}.

\bibitem[{Huang et~al.(2019{\natexlab{a}})Huang, Trabelsi, and
  Za{\"\i}ane}]{huang2019ana}
Chenyang Huang, Amine Trabelsi, and Osmar Za{\"\i}ane. 2019{\natexlab{a}}.
\newblock \href {https://doi.org/10.18653/v1/S19-2006} {{ANA} at
  {S}em{E}val-2019 task 3: Contextual emotion detection in conversations
  through hierarchical {LSTM}s and {BERT}}.
\newblock In \emph{Proceedings of the 13th International Workshop on Semantic
  Evaluation}, pages 49--53, Minneapolis, Minnesota, USA. Association for
  Computational Linguistics.

\bibitem[{Huang et~al.(2019{\natexlab{b}})Huang, Lee, Ma, Chen, Yu, and
  Chen}]{huang2019emotionx}
Yen-Hao Huang, Ssu-Rui Lee, Mau-Yun Ma, Yi-Hsin Chen, Ya-Wen Yu, and Yi-Shin
  Chen. 2019{\natexlab{b}}.
\newblock \href {https://arxiv.org/abs/1908.06264} {Emotionx-idea: Emotion
  bert--an affectional model for conversation}.
\newblock \emph{arXiv preprint arXiv:1908.06264}.

\bibitem[{Jack et~al.(2014)Jack, Garrod, and Schyns}]{jack2014dynamic}
Rachael~E Jack, Oliver~GB Garrod, and Philippe~G Schyns. 2014.
\newblock Dynamic facial expressions of emotion transmit an evolving hierarchy
  of signals over time.
\newblock \emph{Current biology}, 24(2):187--192.

\bibitem[{Jones et~al.(1999)Jones, McCallum, Nigam, and
  Riloff}]{jones1999bootstrapping}
Rosie Jones, Andrew McCallum, Kamal Nigam, and Ellen Riloff. 1999.
\newblock \href
  {https://citeseerx.ist.psu.edu/viewdoc/download?doi=10.1.1.29.2001&rep=rep1&type=pdf}
  {Bootstrapping for text learning tasks}.
\newblock In \emph{IJCAI-99 Workshop on Text Mining: Foundations, Techniques
  and Applications}, volume~1. Citeseer.

\bibitem[{Jou et~al.(2014)Jou, Bhattacharya, and Chang}]{jou2014predicting}
Brendan Jou, Subhabrata Bhattacharya, and Shih-Fu Chang. 2014.
\newblock \href {https://doi.org/10.1145/2647868.2656408} {Predicting viewer
  perceived emotions in animated gifs}.
\newblock In \emph{Proceedings of the 22nd ACM international conference on
  Multimedia}, pages 213--216.

\bibitem[{Katz et~al.(2007)Katz, Singleton, and Wicentowski}]{katz2007swat}
Phil Katz, Matt Singleton, and Richard Wicentowski. 2007.
\newblock \href {https://www.aclweb.org/anthology/S07-1067} {{SWAT}-{MP}:the
  {S}em{E}val-2007 systems for task 5 and task 14}.
\newblock In \emph{Proceedings of the Fourth International Workshop on Semantic
  Evaluations ({S}em{E}val-2007)}, pages 308--313, Prague, Czech Republic.
  Association for Computational Linguistics.

\bibitem[{Krebs et~al.(2017)Krebs, Lubascher, Moers, Schaap, and
  Spanakis}]{krebs2017social}
Florian Krebs, Bruno Lubascher, Tobias Moers, Pieter Schaap, and Gerasimos
  Spanakis. 2017.
\newblock \href {https://arxiv.org/abs/1712.03249} {Social emotion mining
  techniques for facebook posts reaction prediction}.
\newblock \emph{arXiv preprint arXiv:1712.03249}.

\bibitem[{Lei et~al.(2014)Lei, Rao, Li, Quan, and Wenyin}]{lei2014towards}
Jingsheng Lei, Yanghui Rao, Qing Li, Xiaojun Quan, and Liu Wenyin. 2014.
\newblock \href {https://doi.org/https://doi.org/10.1016/j.future.2013.09.024}
  {Towards building a social emotion detection system for online news}.
\newblock \emph{Future Generation Computer Systems}, 37:438 -- 448.
\newblock Special Section: Innovative Methods and Algorithms for Advanced
  Data-Intensive Computing Special Section: Semantics, Intelligent processing
  and services for big data Special Section: Advances in Data-Intensive
  Modelling and Simulation Special Section: Hybrid Intelligence for Growing
  Internet and its Applications.

\bibitem[{Lin et~al.(2008)Lin, Yang, and Chen}]{lin2008emotion}
Kevin Hsin-Yih Lin, Changhua Yang, and Hsin-Hsi Chen. 2008.
\newblock \href {https://doi.org/10.1109/WIIAT.2008.197} {Emotion
  classification of online news articles from the reader's perspective}.
\newblock In \emph{2008 IEEE/WIC/ACM International Conference on Web
  Intelligence and Intelligent Agent Technology}, volume~1, pages 220--226.
  IEEE.

\bibitem[{Ma et~al.(2005)Ma, Prendinger, and Ishizuka}]{ma2005emotion}
Chunling Ma, Helmut Prendinger, and Mitsuru Ishizuka. 2005.
\newblock \href {https://doi.org/10.1007/11573548_80} {Emotion estimation and
  reasoning based on affective textual interaction}.
\newblock In \emph{Proceedings of the First International Conference on
  Affective Computing and Intelligent Interaction}, ACII'05, page 622–628,
  Berlin, Heidelberg. Springer-Verlag.

\bibitem[{Mohammad and Kiritchenko(2015)}]{mohammad2015using}
Saif~M Mohammad and Svetlana Kiritchenko. 2015.
\newblock \href {https://doi.org/10.1111/coin.12024} {Using hashtags to capture
  fine emotion categories from tweets}.
\newblock \emph{Computational Intelligence}, 31(2):301--326.

\bibitem[{Mohammad et~al.(2013)Mohammad, Kiritchenko, and
  Zhu}]{mohammad2013nrc}
Saif~M Mohammad, Svetlana Kiritchenko, and Xiaodan Zhu. 2013.
\newblock Nrc-canada: Building the state-of-the-art in sentiment analysis of
  tweets.
\newblock \emph{arXiv preprint arXiv:1308.6242}.

\bibitem[{Plutchik(1980)}]{plutchik1980general}
Robert Plutchik. 1980.
\newblock \href
  {https://doi.org/https://doi.org/10.1016/B978-0-12-558701-3.50007-7} {Chapter
  1 - a general psychoevolutionary theory of emotion}.
\newblock In Robert Plutchik and Henry Kellerman, editors, \emph{Theories of
  Emotion}, pages 3 -- 33. Academic Press.

\bibitem[{Pool and Nissim(2016)}]{pool2016distant}
Chris Pool and Malvina Nissim. 2016.
\newblock \href {https://www.aclweb.org/anthology/W16-4304} {Distant
  supervision for emotion detection using {F}acebook reactions}.
\newblock In \emph{Proceedings of the Workshop on Computational Modeling of
  People{'}s Opinions, Personality, and Emotions in Social Media ({PEOPLES})},
  pages 30--39, Osaka, Japan. The COLING 2016 Organizing Committee.

\bibitem[{Poria et~al.(2014)Poria, Gelbukh, Cambria, Hussain, and
  Huang}]{poria2014emosenticspace}
Soujanya Poria, Alexander Gelbukh, Erik Cambria, Amir Hussain, and Guang-Bin
  Huang. 2014.
\newblock \href {https://doi.org/https://doi.org/10.1016/j.knosys.2014.06.011}
  {Emosenticspace: A novel framework for affective common-sense reasoning}.
\newblock \emph{Knowledge-Based Systems}, 69:108 -- 123.

\bibitem[{Raad et~al.(2018)Raad, Philipp, Patrick, and
  Christoph}]{raad2018aseds}
Bin~Tareaf Raad, Berger Philipp, Hennig Patrick, and Meinel Christoph. 2018.
\newblock \href {https://doi.org/10.1109/HPCC/SmartCity/DSS.2018.00143} {Aseds:
  Towards automatic social emotion detection system using facebook reactions}.
\newblock In \emph{2018 IEEE 20th International Conference on High Performance
  Computing and Communications; IEEE 16th International Conference on Smart
  City; IEEE 4th International Conference on Data Science and Systems
  (HPCC/SmartCity/DSS)}, pages 860--866. IEEE.

\bibitem[{Rao et~al.(2014)Rao, Li, Wenyin, Wu, and Quan}]{rao2014affective}
Yanghui Rao, Qing Li, Liu Wenyin, Qingyuan Wu, and Xiaojun Quan. 2014.
\newblock \href {https://doi.org/https://doi.org/10.1016/j.neunet.2014.05.007}
  {Affective topic model for social emotion detection}.
\newblock \emph{Neural Networks}, 58:29 -- 37.
\newblock Special Issue on ``Affective Neural Networks and Cognitive Learning
  Systems for Big Data Analysis''.

\bibitem[{Reimers and Gurevych(2019)}]{reimers-2019-sentence-bert}
Nils Reimers and Iryna Gurevych. 2019.
\newblock \href {https://www.aclweb.org/anthology/D19-1410.pdf} {Sentence-bert:
  Sentence embeddings using siamese bert-networks}.
\newblock In \emph{Proceedings of the 2019 Conference on Empirical Methods in
  Natural Language Processing}. Association for Computational Linguistics.

\bibitem[{Rosa et~al.(2015)Rosa, Rodriguez, and Bressan}]{rosa2015music}
Renata~L Rosa, Demsteneso~Z Rodriguez, and Gra{\c{c}}a Bressan. 2015.
\newblock Music recommendation system based on user's sentiments extracted from
  social networks.
\newblock \emph{IEEE Transactions on Consumer Electronics}, 61(3):359--367.

\bibitem[{Rosa et~al.(2018)Rosa, Schwartz, Ruggiero, and
  Rodr{\'\i}guez}]{rosa2018knowledge}
Renata~Lopes Rosa, Gisele~Maria Schwartz, Wilson~Vicente Ruggiero, and
  Dem{\'o}stenes~Zegarra Rodr{\'\i}guez. 2018.
\newblock A knowledge-based recommendation system that includes sentiment
  analysis and deep learning.
\newblock \emph{IEEE Transactions on Industrial Informatics}, 15(4):2124--2135.

\bibitem[{Scherer and Wallbott(1994)}]{scherer1994evidence}
Klaus~R. Scherer and Harald Wallbott. 1994.
\newblock \href {https://doi.org/10.1037//0022-3514.66.2.310} {Evidence for
  universality and cultural variation of differential emotion response
  patterning}.
\newblock \emph{Journal of personality and social psychology}, 66(2):310.

\bibitem[{Sintsova and Pu(2016)}]{sintsova2016dystemo}
Valentina Sintsova and Pearl Pu. 2016.
\newblock \href {https://doi.org/10.1145/2912147} {Dystemo: Distant supervision
  method for multi-category emotion recognition in tweets}.
\newblock \emph{ACM Trans. Intell. Syst. Technol.}, 8(1).

\bibitem[{Staiano and Guerini(2014)}]{staiano2014depechemood}
Jacopo Staiano and Marco Guerini. 2014.
\newblock \href {https://doi.org/10.3115/v1/P14-2070} {Depeche mood: a lexicon
  for emotion analysis from crowd annotated news}.
\newblock In \emph{Proceedings of the 52nd Annual Meeting of the Association
  for Computational Linguistics (Volume 2: Short Papers)}, pages 427--433,
  Baltimore, Maryland. Association for Computational Linguistics.

\bibitem[{Stark and Hoey(2021)}]{stark2021ethics}
Luke Stark and Jesse Hoey. 2021.
\newblock \href {https://doi.org/10.1145/3442188.3445939} {The ethics of
  emotion in artificial intelligence systems}.
\newblock In \emph{Proceedings of the 2021 ACM Conference on Fairness,
  Accountability, and Transparency}, pages 782--793.

\bibitem[{Strapparava and Mihalcea(2007)}]{strapparava2007semeval}
Carlo Strapparava and Rada Mihalcea. 2007.
\newblock Semeval-2007 task 14: Affective text.
\newblock \emph{Association for Computational Linguistics}, pages 308--313.

\bibitem[{Strapparava and Valitutti(2004)}]{strapparava2004wordnet}
Carlo Strapparava and Alessandro Valitutti. 2004.
\newblock \href {http://www.lrec-conf.org/proceedings/lrec2004/pdf/369.pdf}
  {{W}ord{N}et affect: an affective extension of {W}ord{N}et}.
\newblock In \emph{Proceedings of the Fourth International Conference on
  Language Resources and Evaluation ({LREC}{'}04)}, Lisbon, Portugal. European
  Language Resources Association (ELRA).

\bibitem[{Suttles and Ide(2013)}]{suttles2013distant}
Jared Suttles and Nancy Ide. 2013.
\newblock \href {https://doi.org/10.1007/978-3-642-37256-8_11} {Distant
  supervision for emotion classification with discrete binary values}.
\newblock In \emph{International Conference on Intelligent Text Processing and
  Computational Linguistics}, pages 121--136. Springer.

\bibitem[{Tao(2004)}]{tao2004context}
Jianhua Tao. 2004.
\newblock \href {http://www.speakit.cn/Group/file/Emotion04.pdf} {Context based
  emotion detection from text input}.
\newblock In \emph{Eighth International Conference on Spoken Language
  Processing}.

\bibitem[{Tapia()}]{tapia}
Luis Tapia.
\newblock \href
  {https://www.shutterstock.com/image-photo/clown-on-black-background-315814736}
  {\emph{Clown on black background}}.
\newblock Shutterstock.

\bibitem[{Tkal{\v{c}}i{\v{c}} et~al.(2017)Tkal{\v{c}}i{\v{c}}, Maleki, Pesek,
  Elahi, Ricci, and Marolt}]{tkalvcivc2017research}
Marko Tkal{\v{c}}i{\v{c}}, Nima Maleki, Matev{\v{z}} Pesek, Mehdi Elahi,
  Francesco Ricci, and Matija Marolt. 2017.
\newblock A research tool for user preferences elicitation with facial
  expressions.
\newblock In \emph{Proceedings of the eleventh acm conference on recommender
  systems}, pages 353--354.

\bibitem[{Tkal{\v{c}}i{\v{c}} et~al.(2019)Tkal{\v{c}}i{\v{c}}, Maleki, Pesek,
  Elahi, Ricci, and Marolt}]{tkalvcivc2019prediction}
Marko Tkal{\v{c}}i{\v{c}}, Nima Maleki, Matev{\v{z}} Pesek, Mehdi Elahi,
  Francesco Ricci, and Matija Marolt. 2019.
\newblock Prediction of music pairwise preferences from facial expressions.
\newblock In \emph{Proceedings of the 24th International Conference on
  Intelligent User Interfaces}, pages 150--159.

\bibitem[{Wang et~al.(2011)Wang, Guo, Wang, and Wang}]{wang2011drosophila}
Kaiyu Wang, Yanmeng Guo, Fei Wang, and Zuoren Wang. 2011.
\newblock Drosophila trpa channel painless inhibits male--male courtship
  behavior through modulating olfactory sensation.
\newblock \emph{PLoS One}, 6(11):e25890.

\bibitem[{Wang et~al.(2012)Wang, Chen, Thirunarayan, and
  Sheth}]{wang2012harnessing}
Wenbo Wang, Lu~Chen, Krishnaprasad Thirunarayan, and Amit~P Sheth. 2012.
\newblock \href {https://doi.org/10.1109/SocialCom-PASSAT.2012.119} {Harnessing
  twitter "big data" for automatic emotion identification}.
\newblock In \emph{2012 International Conference on Privacy, Security, Risk and
  Trust and 2012 International Confernece on Social Computing}, pages 587--592.
  IEEE.

\bibitem[{Wolf et~al.(2020)Wolf, Debut, Sanh, Chaumond, Delangue, Moi, Cistac,
  Rault, Louf, Funtowicz, Davison, Shleifer, von Platen, Ma, Jernite, Plu, Xu,
  Le~Scao, Gugger, Drame, Lhoest, and Rush}]{wolf2019huggingface}
Thomas Wolf, Lysandre Debut, Victor Sanh, Julien Chaumond, Clement Delangue,
  Anthony Moi, Pierric Cistac, Tim Rault, Remi Louf, Morgan Funtowicz, Joe
  Davison, Sam Shleifer, Patrick von Platen, Clara Ma, Yacine Jernite, Julien
  Plu, Canwen Xu, Teven Le~Scao, Sylvain Gugger, Mariama Drame, Quentin Lhoest,
  and Alexander Rush. 2020.
\newblock \href {https://doi.org/10.18653/v1/2020.emnlp-demos.6} {Transformers:
  State-of-the-art natural language processing}.
\newblock In \emph{Proceedings of the 2020 Conference on Empirical Methods in
  Natural Language Processing: System Demonstrations}, pages 38--45, Online.
  Association for Computational Linguistics.

\bibitem[{Zheng et~al.(2016)Zheng, Gu, Lei, Lu, Wang, Li, and
  Wang}]{zheng2016safety}
Zheng Zheng, Simeng Gu, Yu~Lei, Shanshan Lu, Wei Wang, Yang Li, and Fushun
  Wang. 2016.
\newblock Safety needs mediate stressful events induced mental disorders.
\newblock \emph{Neural plasticity}, 2016.

\end{thebibliography}
\bibliographystyle{acl_natbib}

\clearpage

\appendix

\setcounter{figure}{0}  
\setcounter{table}{0}
\renewcommand\thefigure{\Alph{section}\arabic{figure}} 
\renewcommand{\thetable}{\Alph{section}\arabic{table}}

\section{Broader Impact}

Any work that touches upon emotion recognition or recognizing affective response needs to ensure that it is sensitive to the various ways of expressing affect in different cultures and individuals. Clearly, applying the ideas described in this paper in a production setting would have to first test for cultural biases. To make ``broad assumptions about emotional universalism [would be] not just unwise, but actively deleterious'' to the general community \citep{stark2021ethics}. We also note that emotion recognition methods belong to a taxonomy of conceptual models for emotion (such as that of \citet{stark2021ethics} and these ``paradigms for human emotions [...] should [not] be taken naively ground truth.''

Before being put in production, the method would also need to be re-evaluated when applied to a new domain to ensure reliable performance in order to prevent unintended consequences. Additionally, our work in detecting affective response is intended for understanding content, not the emotional state of individuals. This work is intended to identify or recommend content, which aligns with the user's preferences. This work should not be used for ill-intended purposes such as purposefully recommending particular content to manipulate a user's perception or preferences.

\section{Details on expanding CARE}
\label{sec:ngram_freq}

The six CARE patterns that were used to seed the system are the following: (The symbol $^*$ (resp.~$^+$) indicates that zero (resp.\ one) or more matches are required.) Example: \textit{This is so amazing!}

\begin{itemize}
    \item Demonstrative Pronouns:\\ \{this$\vert$that$\vert$those$\vert$these\}\{is$\vert$are\}*\{$E$\}$^*$\{$I$\}$^+$\\
    Example: \textit{This is so amazing!}
    \item Subjective Self Pronouns:\\ \{i$\vert$we\}\{am$\vert$is$\vert$are$\vert$have$\vert$has\}*\{$E$\}$^*$\{$I$\}$^+$\\
    Example: \textit{I am really inspired by this recipe.}
    \item Subjective Non-self Pronouns:\\
    \{he$\vert$she$\vert$they\}\{is$\vert$are$\vert$have$\vert$has\}*\{$E$\}$^*$\{$I$\}$^+$\\
    Example: \textit{They really make me mad.}
    \item Collective Nouns:\\ 
    \{some people$\vert$humans$\vert$society\}\{${E}$\}$^+$\{$I$\}$^+$\\
    Example: \textit{Some people are so dumb.}
    \item Leading Exaggerators: \{${E}$\}$^+$\{$I$\}$^+$\\
    Example: \textit{So sad to see this still happens.}
    \item Exclamatory Interrogatives: \\\{what a$\vert$how\}\{${E}$\}$^+$\{$I$\}$^+$\\
    Example: \textit{What a beautiful baby!}
\end{itemize}

Algorithm~\ref{alg:expand} on page~\pageref{alg:expand} presents pseudo-code for the process of labeling posts and expanding CARE patterns and the CARE lexicon. Table~\ref{tab:ngram_examples} presents example results from the expansion process.

\begin{table}[H]
\small
\centering
\begin{tabular}{ ccc } 
 \toprule
 \textbf{n-gram} & \textbf{frequency} & \textbf{class} \\
 \midrule
adorable & 9000 & Adoring \\
gorgeous & 8422 & Adoring \\
fantastic & 7796 & Approving \\
interesting & 5742 & Amused\\
sorry for your & 5202 & Saddened\\
brilliant & 4205 & Approving \\
fake & 2568 & Angered \\
sorry to hear & 2323 & Saddened \\
why i hate & 1125 & Angered \\
\midrule
i feel like & 293 & pattern \\
you are a & 207 & pattern \\
this is the & 173 & pattern \\
this made me & 110 & pattern \\
he is so & 102 & pattern \\
\bottomrule
\end{tabular}
\caption{Examples of n-grams resulting from \texttt{GetNgrams} in Algorithm~\ref{alg:expand} and steps B1 and B2 of Figure~\ref{fig:care_diagram}. The n-grams above the middle line are added to the lexicon under the specific class listed while the n-grams below are used for further expansion of CARE patterns after translating to reg-ex format manually.}
\label{tab:ngram_examples}
\end{table}

\label{sec:alg_details}
\begin{algorithm*}[h]
    \caption{Algorithm for producing candidates for new CARE patterns and indicators in the CARE lexicon. Algorithm uses three hyperparameters $t$ (the minimum number of comments to label a post), $f\_lexicon$ (the minimum frequency of a ngram to be added to the lexicon), and $f\_pattern$ (the minimum frequency of an ngram to be a candidate pattern) which was set to 5, 1000, and 100, respectively. The resulting list of candidate patterns needs to be manually converted into a regular expression matching the structure outlined in Section \ref{sec:care-patterns}.}
    \label{alg:expand}
    \KwData{$C$: set of comments, $P$: set of corresponding posts, $L$: dictionary of keywords to class (CARE lexicon), $D$: list of non-class-specific regular expressions (CARE patterns)}
    \DontPrintSemicolon
    
    \SetKwFunction{posts}{LabelPosts}
    \SetKwFunction{ngrams}{GetNgrams}
    
    $lexicon\_candidates \gets []$, $pattern\_candidates \gets []$\;
    $labeled\_posts$ $\gets$ \posts($C$, $P$, $L$, $D$), $ngrams$ $\gets$ \ngrams(labeled\_posts, $C$)\;
    
    \For{a in all classes}{
     // Add an ngram as a lexicon candidate if it is exclusively in high frequency with class $a$\;
        \For{ngram in ngrams[a]}{
            \If{frequency of ngram in ngrams[a] $\geq f\_lexicon$}{
                \For{b in all classes where b $\neq$ a}{
                    \If{ngram in ngrams[b] and frequency of ngram in ngrams[b] $\geq f\_lexicon$}{
                        Break and continue to new n-gram
                    }
                }
                Append ngram to $lexicon\_candidates$, if not added already
            }
        }
    // Add an ngram as a pattern candidate if in high enough frequency and present in another class\;
    \For{ngram in ngrams[a]}{
        \If{total freq. of ngram in ngrams $\geq$  $f\_pattern$ and ngram in ngrams[b] for $b \neq a$}{
                Append $ngram$ to $pattern\_candidates$, if not added already\;
        }
    }
    }

   \KwResult{$lexicon\_candidates$, $pattern\_candidates$}\;
 
    \SetKwProg{Fn}{Function}{:}{}
    \Fn{\posts{$C$, $P$, $L$, $D$}}{
        $labeled\_comments \gets \{\}$, $labeled\_posts \gets \{\}$\;
        
        // For each comment, apply reg-ex and map indicator to affect using the lexicon \;
        \For{c in C}{  
            \If{indicator is non-empty after reg-ex matching and in lexicon}
            {
                Append $c$ to $labeled\_comments[L[indicator]]$\; 
            }
        }
        
         // For each post, aggregate comment labels to label post \;
        \For{$p$ in $P$}{   
            \For{$a$ in all classes}{
                \If{number of comments belonging to post $p$ and labeled as class $a \geq t$ }{
                    Append $p$ to $labeled\_posts[a]$\;
                }
            }
        }
        \KwRet $labeled\_posts$\;
    }\;
    
    \SetKwProg{Fn}{Function}{:}{}
    \Fn{\ngrams{$labeled\_posts$, $C$}}{
        $ngrams \gets \{\}$\;
        \For{$a$ in all classes}{
            // Get the n-grams of all comments belonging to a post labeled as $a$\;
            \For{$p$ in labeled\_posts[$a$]}{
                \For{c in C belonging to post p}{
                    Add 1-grams, 2-grams, and 3-grams of comment $c$ to $ngrams[a]$
                }
            }
        }
        
        \KwRet $ngrams$\;
    }
\end{algorithm*}

\section{Annotation details}

\begin{figure}[H]
  \centering
  \includegraphics[width=0.7\linewidth]{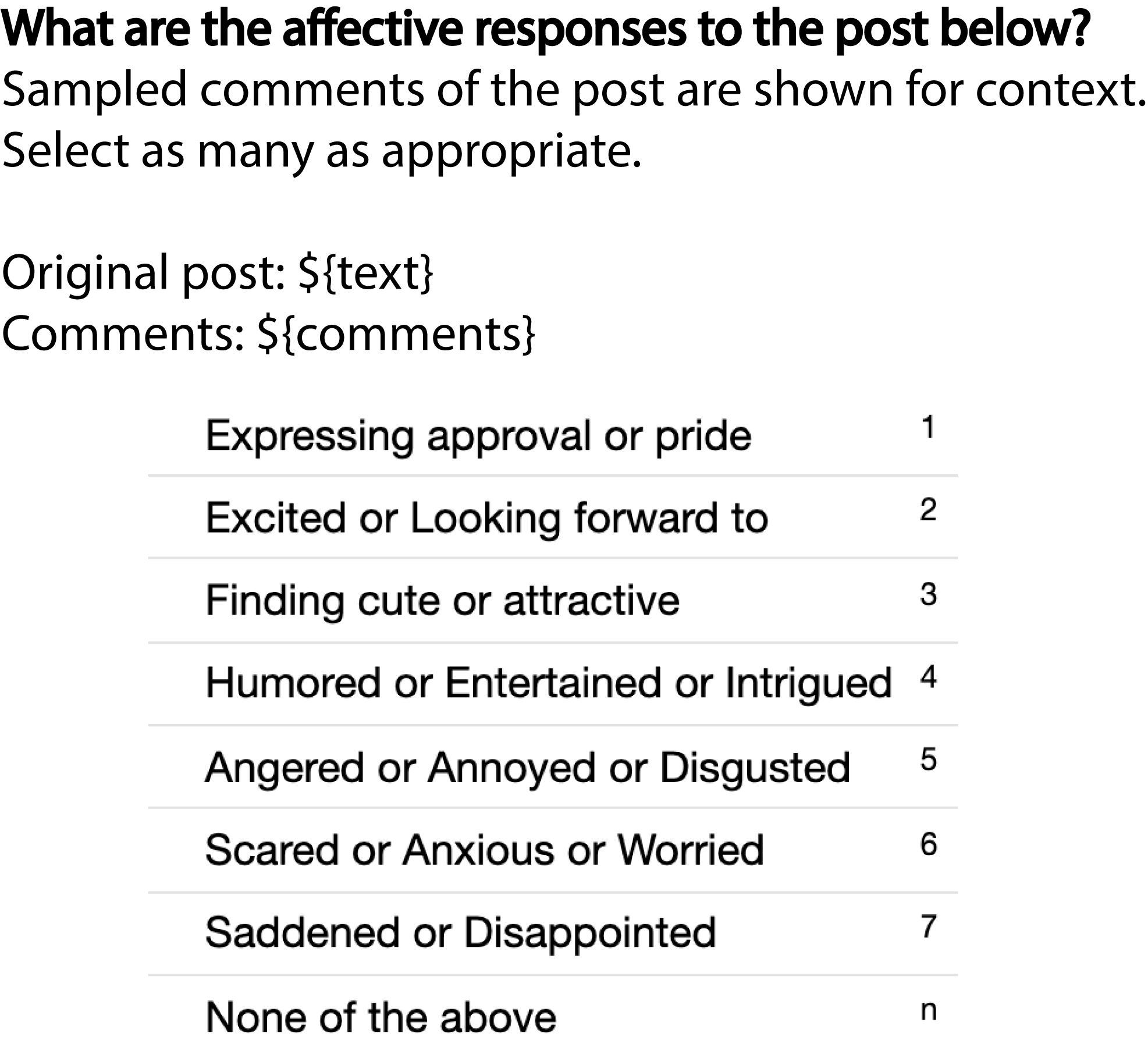} 
  \caption{Interface for crowd-sourcing process using Amazon Mechanical Turk. Three distinct annotators were used to annotate each post. Annotators were told an affective response is an emotion or cognitive response to the post and the definitions and examples in Table~\ref{tab:care_definitions} were shown to them.}
  \label{fig:amt_ui}
\end{figure}

Figure~\ref{fig:amt_ui} shows the interface used for crowd-sourcing human annotations for evaluating CARE patterns. To better understand annotation results for each class, we present Table~\ref{tab:amt_class_results}, which shows annotator agreement statistics broken down by class. We also computed Fleiss' kappa for each class, where a value between 0.41-0.60 is generally considered moderate agreement and a value between 0.61-0.80 is substantial agreement. As can be seen, classes such as \textit{adoring} have high average annotator support and Fleiss' kappa while others like \textit{amused} have low average annotator support and Fleiss' kappa, an observation that aligns with the findings in Section~\ref{sec:error_analysis}.

\begin{table}[H]
\centering
\small
\begin{tabular}{cccc} 
\toprule
 \textbf{AR} & \textbf{\% w/} & \textbf{Avg}  & \textbf{Fleiss'}\\
 & \textbf{support} & \textbf{support} & \textbf{kappa} \\
\midrule
  Adoring & 99.2 & 2.8 & 0.78\\
  Amused & 93.2 & 2.1 & 0.43\\
  Approving & 98.8 & 2.8 & 0.51\\
  Excited & 83.6 & 2.1 & 0.58\\
  Angered & 99.4 & 2.8 & 0.59\\
  Saddened  & 99.6 & 2.9 & 0.61\\
  Scared & 98.8 & 2.6 & 0.64\\
  \textbf{Average}& \textbf{96.1}  & \textbf{2.6} & \textbf{0.59}\\
  \bottomrule
  \hline
\end{tabular}
\caption{The percent of CARE-labeled examples (maximum of 100) with agreement from at least one labeler by class and of those examples, the average number of annotator agreement (maximum of 3). The third column shows the Fleiss' kappa, which was computed for class $a$ based on the presence and absence of label $a$ by each annotator for a given post. The bottom row is the average over all classes.}
\label{tab:amt_class_results}
\end{table}


\begin{figure}[H]
  \centering
  \includegraphics[height=6.2cm]{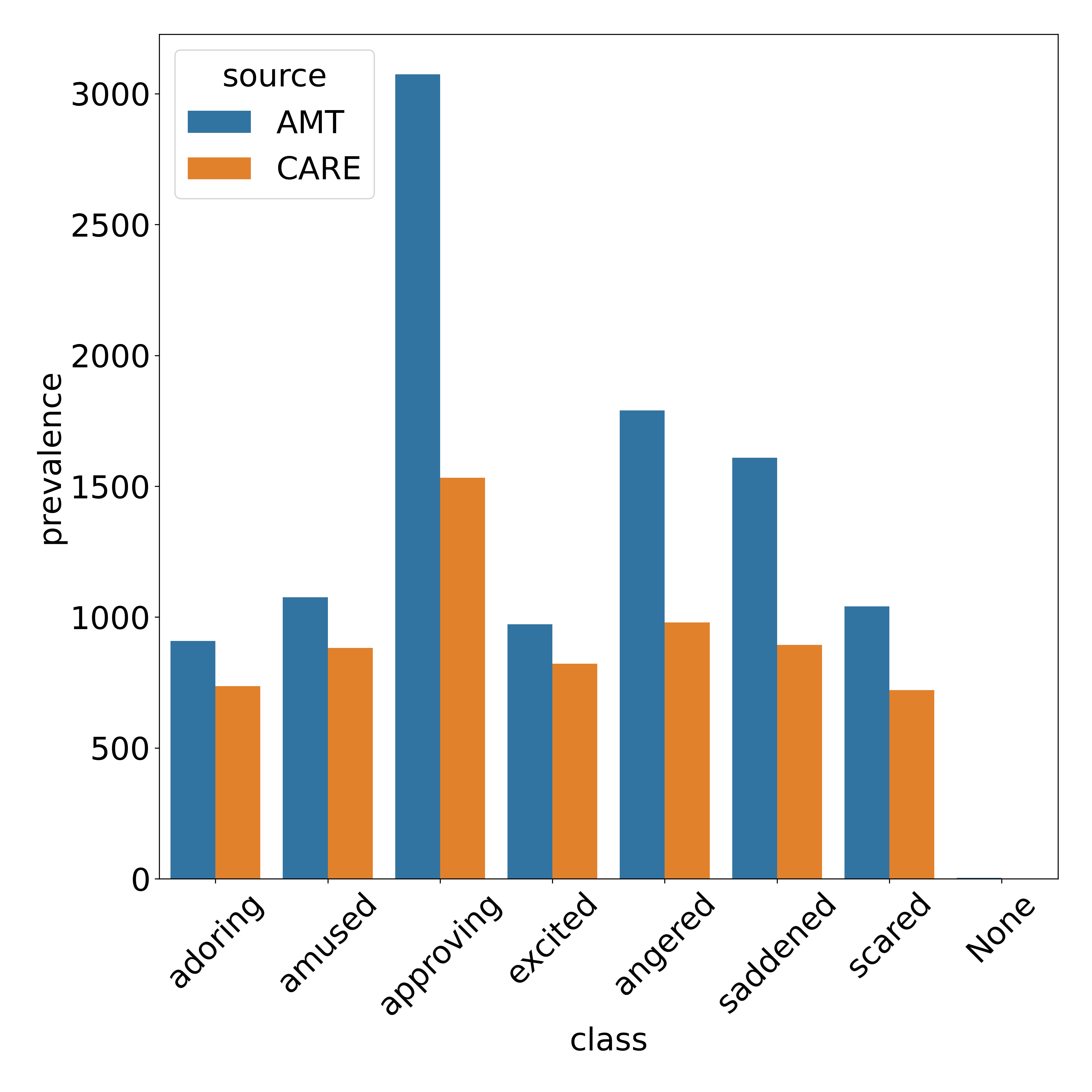} 
  \caption{Prevalence of class labels according to annotations from AMT on which at least two annotators agree upon (blue) and according to CARE (orange). The prevalence of \textit{approving} was much higher from AMT, likely due to a large perceived overlap in the definitions of \textit{approving} and other classes such \textit{excited}.}
  \label{fig:class_prevalence}
\end{figure}


\begin{figure}[H]
  \centering
  \includegraphics[width=0.85\linewidth]{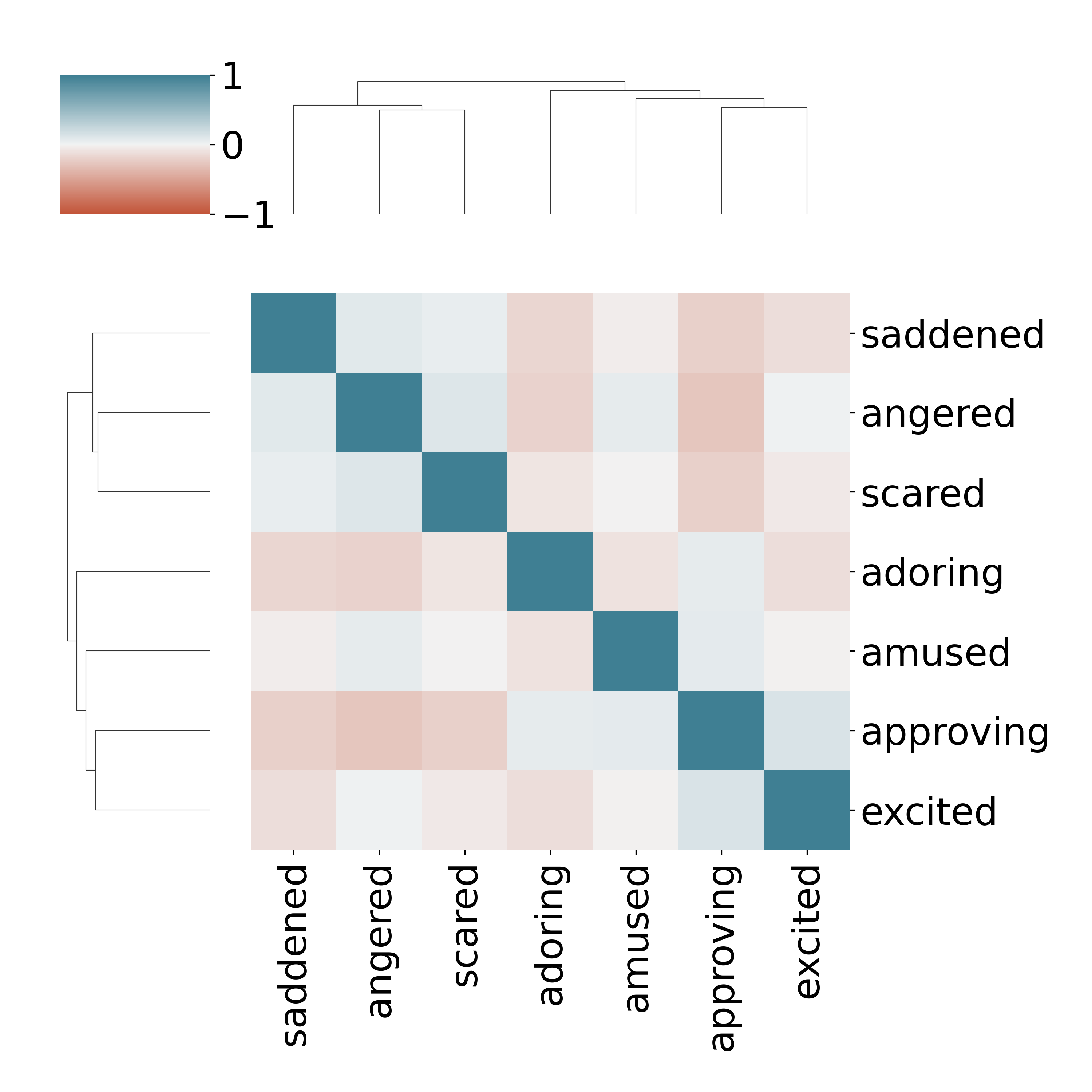} 
  \caption{Pairwise Spearman correlation between each pair of classes, computed using the degree of annotator support for each class given a post. The dendrogram represents a hierarchical clustering of the data, correctly capturing the distinction between positive and negative classes.}
  \label{fig:corr_heatmap}
\end{figure}

\section{Are affective response and publisher affect the same?}

\label{sec:response_and_publisher}

\begin{figure}[H]
\centering
  \includegraphics[width=\linewidth]{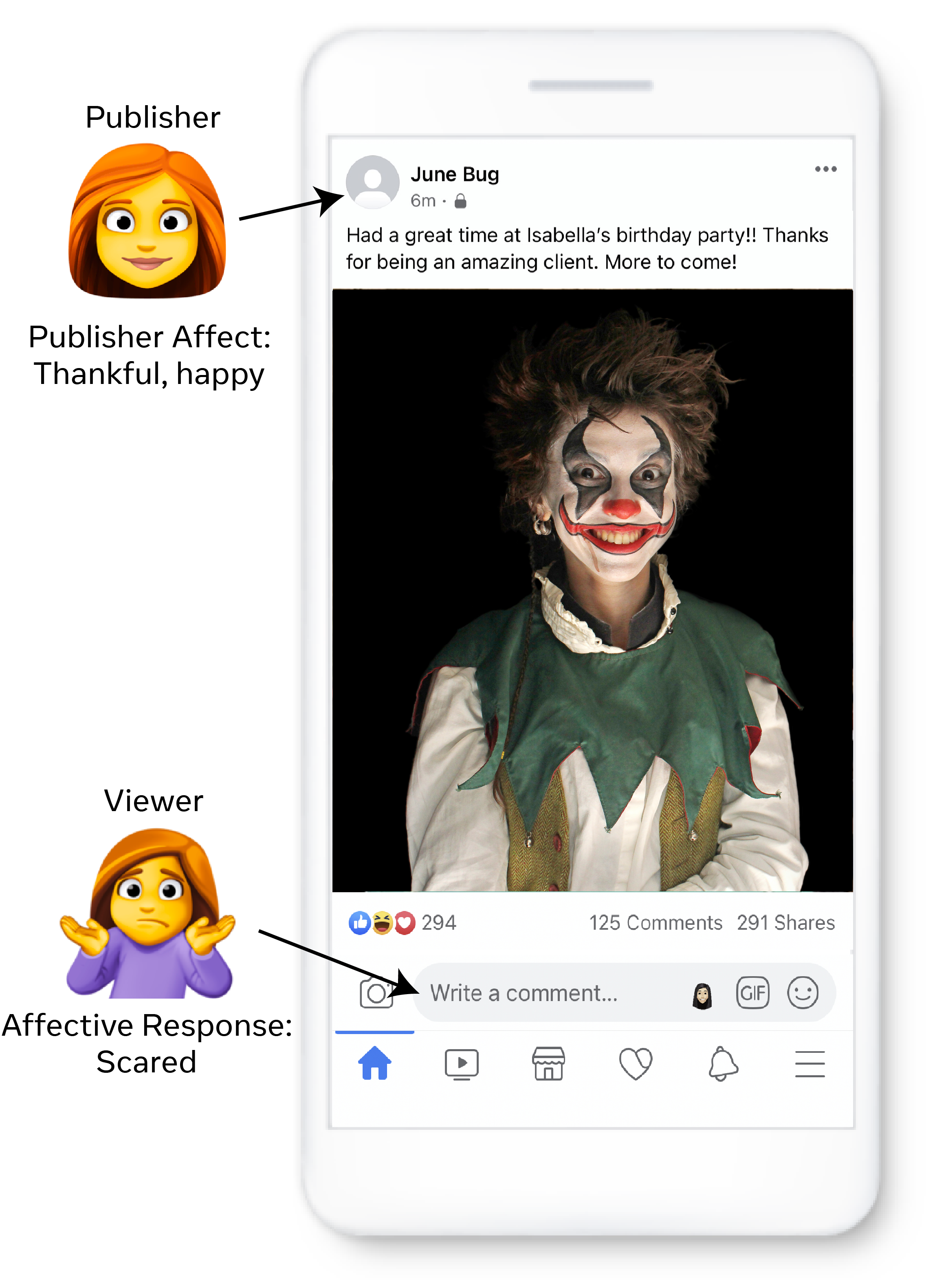}
  \caption{An example case of differing publisher affect and affective response. This work focuses on affective response through signals such as comments and reactions. Post image sourced from Shutterstock \cite{tapia}.}
  \label{fig:clown}
\end{figure}

The GoEmotions dataset and classifier target the publisher affect (of comments), whereas \cbert\ and CARE target the affective response (of posts). In an effort to study the correlation between affective response and publisher affect, we compare the following sets of labels: 1) human annotations of GoEmotion and the predicted affective responses using \cbert~applied to GoEmotions and 2) CARE labels for posts in \cdb~and the predicted publisher affects using the GoEmotions classifier applied to \cdb. Specifically, for every annotated label (i.e., not from a classifier) we count the percentage of the time where there is intersection with the set of predicted labels (i.e., from a classifier).

The results of these experiments are shown in Table~\ref{tab:response_publisher}, broken down according to the class of the annotated label. Overall, the percentage of affective response and publisher affect label agreement (44\%) is moderate but seems to indicate that the affective response detection and emotion detection are not necessarily the same problem, in particular for \textit{scared}  and \textit{approving}.  The classes \textit{approving}, \textit{excited}, and \textit{angered} have a large variance between the two datasets, where the first (Table~\ref{tab:response_publisher}, second column) uses comments and the second (Table~\ref{tab:response_publisher}, third column) uses posts. This could be due to the classification errors (either by GoEmotions or by \cbert) or due to the type of the text (comment or post). More research and data collection is needed to understand the relationship between affective response and publisher affect, which as demonstrated by Figure~\ref{fig:clown} are not necessarily equivalent. 

\begin{table}[H]
\centering
\small
\begin{tabular}{ccccc} 
\toprule
\textbf{AR} &\textbf{GoEmotions} & \textbf{\cdb} & \textbf{Average} \\
\midrule
  Amused & 63 & 54 & 59\\
  Approving & 8 & 47 & 28\\
  Excited & 52 & 24 & 38\\
  Angered & 4 & 74 & 39\\
  Saddened  & 60 & 62 & 61 \\
  Scared & 44 & 34 & 39\\
  \textbf{Average}& \textbf{39}  & \textbf{49} & \textbf{44}\\
  \bottomrule
\end{tabular}
\caption{Rate of intersection between affective response and publisher affect labels. The first column denotes the class. The second column denotes the percent of the time an annotated label in GoEmotions exists in the set of predicted labels by \cbert~when applied to the GoEmotions dataset. The third column denotes the percent of the time an annotated label in \cbert~exists in the set of predicted labels by the GoEmotions classifier when applied to \cdb. The last column is the row-wise average.}
\label{tab:response_publisher}
\end{table}

\section{Using CARE patterns/lexicon to predict publisher affect in GoEmotions}
\label{sec:goemotions}
The GoEmotions dataset~\cite{demszky2020goemotions} is a collection of 58k Reddit comments labeled according to the publisher affect from a taxonomy of 28 emotions. There exists a natural mapping from 6 of our classes to those of GoEmotions (the exception being \textit{adoring}) based on the definitions alone. Hence, applying CARE patterns/lexicon to the GoEmotions dataset presents another way of validating the quality of steps 1 and 2 of CARE. The number of examples in GoEmotions with labels belonging to these 6 classes was 21.0k and the number of comments that were labeled by CARE patterns/lexicon was 1259. Table~\ref{tab:goemotions_class_results} compares the human annotations in the GoEmotions dataset with the labels that CARE patterns/lexicon assigned to the comments and shows that they have a high degree of agreement.

While the low recall is certainly a limitation of CARE patterns and lexicon when applied to a specific small dataset, we emphasize that the primary intention of CARE patterns is to generate a labeled dataset in an unsupervised manner so one can start training classifiers for that affective response. Given the abundance of freely available unlabeled data (e.g., on Reddit, Twitter), recall is not a problem in practice.  In the next section and in Section~\ref{sec:carego}, however, we discuss how existing emotion classifiers, such as the GoEmotions classifier \cite{demszky2020goemotions} can also be leveraged in the CARE method.

\section{CARE and \carego~ evaluation details}
\label{sec:care_g}

\carego~refers to the CARE method where steps 1 and 2 of Figure~\ref{fig:care_diagram} use the GoEmotions classifier instead of CARE patterns. To evaluate how CARE and \carego~compares, we use the same human-labeled dataset described in Section~\ref{sec:annotation} and applied the GoEmotions classifier to all the comments belonging to these posts (72k comments). We then mapped the predicted GoEmotion labels to CARE pattern labels using the mapping in Table~\ref{tab:goemotions_class_results}. GoEmotion and CARE labels not in the mapping are excluded from this analysis. 

The same metrics for $\geq2$ annotator agreement in Table \ref{tab:amt_results} are shown in Table~\ref{tab:amt_results_compare} for multiple thresholds and for all classes, excluding \textit{adoring}. CARE labels consistently demonstrate higher agreement with human annotations than those of \carego. The last row of Table~\ref{tab:amt_results_compare} shows results for an ensembling approach where steps 1 and 2 use labels from both CARE patterns in addition to the labels from the GoEmotions classifier, where the former uses $t=5$ and the latter uses $t=4$ in step 3 (optimal values for each approach, respectively). This ensembling approach does reasonably well and can be used to include classes in the GoEmotions taxonomy that do not exist in the taxonomy of Table~\ref{tab:care_definitions}. Given other emotion classifiers, one could potentially include those as well.

\begin{table}[H]
\small
\centering
\begin{tabular}{cccc} 
\toprule
 \textbf{Threshold} & \textbf{Any \carego} & \textbf{All \carego} & \textbf{Other} \\
\midrule
  $t=1$ & 95 & 34 & 25\\
  $t=2$ & 91 & 61 & 42\\
  $t=3$ & 87 & 71 & 51\\
  $t=4$ & 81 & 73 & 57\\
  $t=5$ & 73 & 67 & 62\\
  $t=6$ & 58 & 56 & 70\\
  $t=7$ & 47 & 45 & 76\\
  $t=8$ & 38 & 37 & 81\\
  $t=9$ & 30 & 29 & 84\\
  $t=10$ & 24 & 23 & 88\\
  max & 89 & 89 & 60\\
  CARE & 93 & 89 & 54 \\
  ensemble & 94 & 83 & 49 \\
  \bottomrule
  \hline
\end{tabular}
\caption{The rate of intersection between labels agreed upon by at least two annotators and the labels proposed by \carego.  The first column indicates the threshold $t$ used in \carego. Using annotations agreed upon by at least two annotators, the rest of the columns show the rate of agreement with at least one predicted label, all predicted labels, and any human-annotated label that was not predicted. The row labeled `max' refers to choosing the comment-level label with the highest frequency for each post. For context, the results for CARE using $t=5$ are shown in the penultimate row. The last row presents results from combining the CARE pattern labels and the GoEmotion labels using $t=4$.}
\label{tab:amt_results_compare}
\end{table}

\section{Multi-dimensional scaling pairwise plots}

\begin{figure}[t]
  \centering
  \small
  \includegraphics[width=\linewidth]{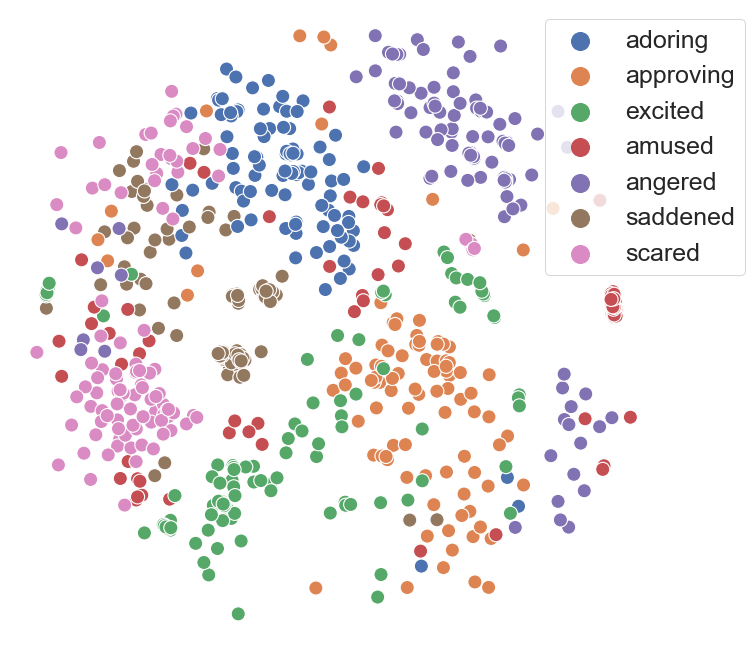} 
  \caption{The two-dimensional projection (using MDS) of sentence embeddings of comments suggests that the CARE-based predictions correspond to similarity in the embedding space.
  Colors correspond to the labels given by CARE labeling, which were not given to the embedding model or the MDS. }
  \label{fig:mds_plot}
\end{figure}

\begin{figure*}[ht]
  \centering
  \includegraphics[width=1\linewidth]{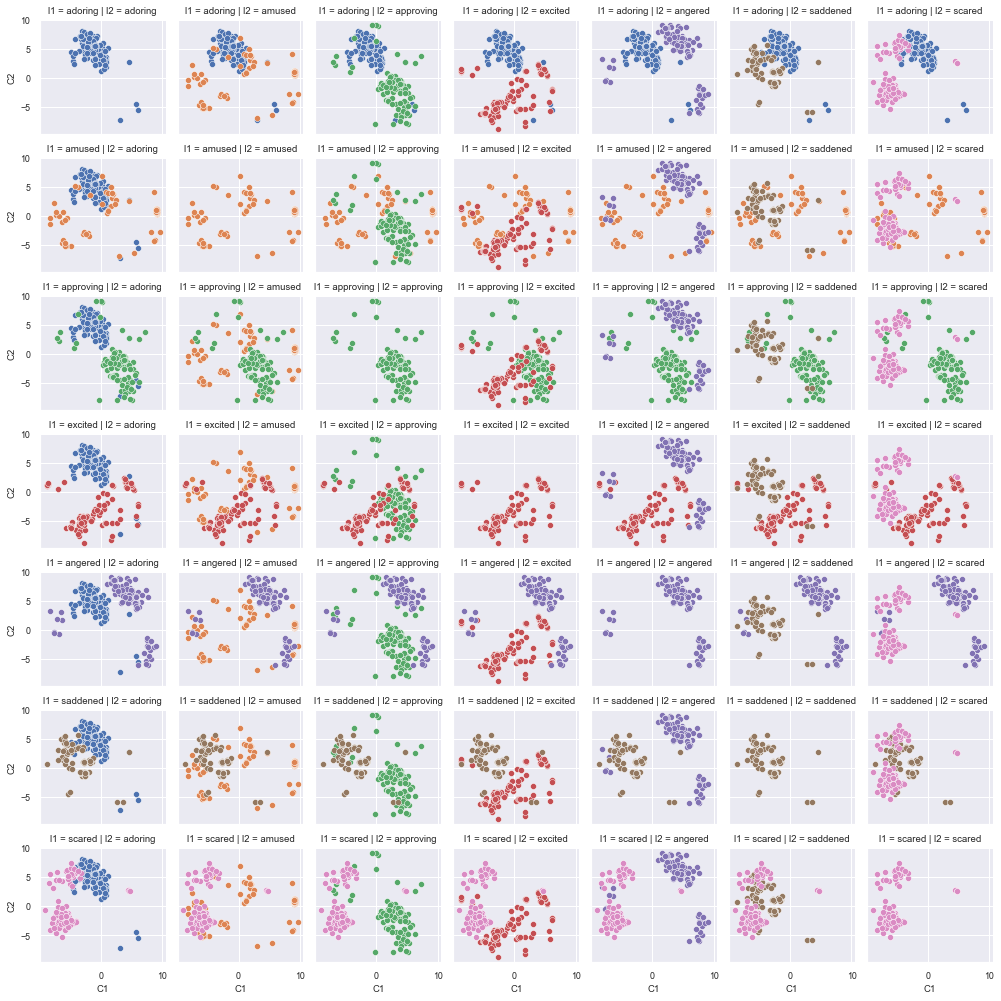} 
  \caption{Subplots of plotting the multi-dimensional scaling from Figure~\ref{fig:mds_plot} for each pairwise comparison of the 7 classes. The rows and columns follow in the order adoring, amused, approving, excited, angered, saddened, and scared. The entire grid is symmetric for ease of exploration. }
  \label{fig:mds_plot_full}
\end{figure*}

We visualize the degree of overlap between the sentence embeddings (using Sentence-Bert \citep{reimers-2019-sentence-bert}) of 100 comments in \cdb~for each class. We then use multi-dimensional scaling or MDS \cite{cox2008multidimensional} to map the embeddings to the same two-dimensional space using euclidean distance as the similarity metric, as shown in Figure~\ref{fig:mds_plot} and Figure~\ref{fig:mds_plot_full}. Note that the MDS process does not use the class labels. As can be seen, there is substantial overlap between \textit{amused} and other classes as well as between \textit{excited} and \textit{approving}. Given that the average number of human annotations per post was 1.8 (Section~\ref{sec:annotation}), it is likely that a portion of this overlap can attributed to the multi-label nature of the problem as well as the moderate correlation between certain classes such as \textit{excited} and \textit{approving} (Figure~\ref{fig:corr_heatmap}). See Figure~\ref{fig:mds_plot_full} for plots of multi-dimensional scaling for every pair of classes, as referenced in Section~\ref{sec:error_analysis}. 

\section{Pattern match analysis}

To investigate why higher thresholds would be needed for certain classes, we analyze the CARE patterns and lexicon at the class level.

Let us define a match as a tuple containing the pattern name and the word or phrase which maps the comment to an affect according to the CARE lexicon. We could also consider exaggerators in our analysis but here we assume a negligible effect on differentiating reliability. We previously assumed that each instantiated match should have the same weight of 1, but this may not be appropriate considering that some patterns or words may be more reliable. 

\begin{figure}[H]
  \centering
  \includegraphics[width=1\linewidth]{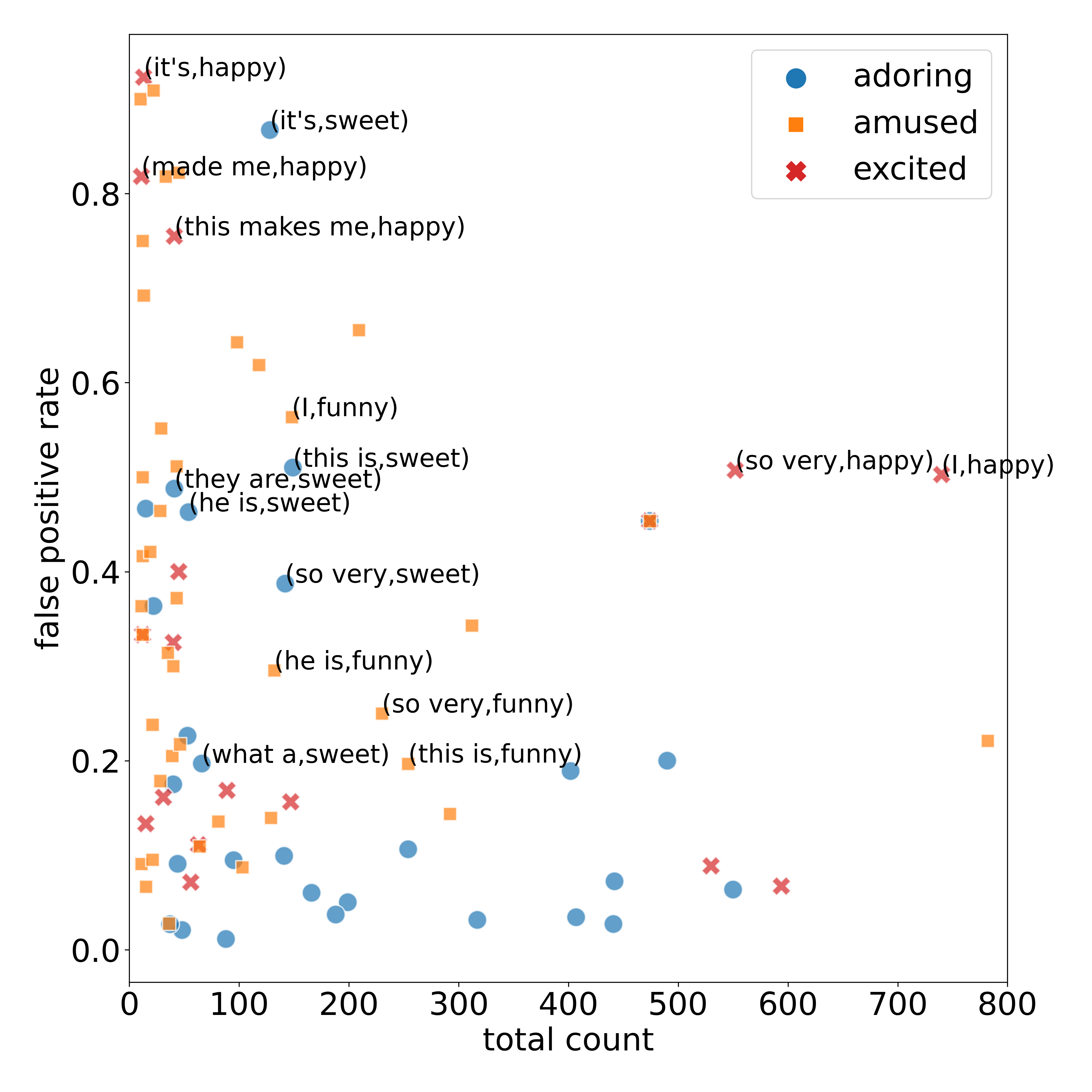} 
  \caption{Scatter plot of the total frequency of a match versus its false positive rate. Ground truth labels used here are those from AMT and agreed upon by at least 2 annotators. For clarity, a match is shown only if its total count was 10 or more and if it belongs to one of the three classes (\textit{adoring}, \textit{amused}, and \textit{excited}). Only those which contain the keywords `sweet' (\textit{adoring}), `funny' (\textit{amused}), and `happy' (\textit{excited}) are labeled.}
  \label{fig:match_analysis}
\end{figure}

As can be seen in Figure~\ref{fig:match_analysis}, there are some cases in which the keyword in general seems to have a high false positive rate (e.g., happy) and in other cases it appears the erroneous combination of a particular pattern and keyword can lead to high false positive rates. For example, while the match `(so very, funny)' has a low false positive rate of 0.2, `(I, funny)' has a much higher false positive rate of 0.57, which intuitively makes since `I'm funny' does not indicate being amused. We also investigated whether individual patterns are prone to higher false positive rates, which does not seem to be the case. For future iterations of CARE, one could also use the true positive rate as the weight of a match to obtain a weighted sum when aggregating over comments to label a post.

\section{Modeling details}
\label{sec:bert_setup_details}

\begin{table}[H]
\centering
\small
\begin{tabular}{ccccc} 
\toprule
\textbf{AR} &\textbf{Precision} & \textbf{Recall} & \textbf{F1} \\
\midrule
  Positive & 0.95 & 0.95 & 0.94 \\
  Negative & 0.77 & 0.77 & 0.78 \\
  \textbf{micro-avg} & \textbf{0.89} & \textbf{0.91} & \textbf{0.90}\\
  \textbf{macro-avg} & \textbf{0.86} & \textbf{0.86} & \textbf{0.86}\\
  \textbf{stdev} & \textbf{0.10} & \textbf{0.13} & \textbf{0.11} \\
  \bottomrule

\end{tabular}
\caption{Accuracy of \cbert~for the two-class case: {\sc positive} versus {\sc negative}. Note that amused, excited, adoring, and approving were mapped to positive and angered, saddened, and scared were mapped to negative.}
\label{tab:pos_neg_bert_results}
\end{table}

We began with the hyper-parameter settings in \citet{demszky2020goemotions} and explored other hyper-parameter settings (batch sizes [16, 32, 64], max length [64, 256, 512], drop out rate [0.3, 0.5, 0.7], epochs [2-10]) but found minimal improvements in the F1-score, as computed by the \texttt{scikit-learn} package in python. Running this on two Tesla P100-SXM2-16GB GPUs took roughly 19 hours. We also experimented with higher thresholds for the parameter $t$ (see Section~\ref{sec:post-labeling}) but saw marginal improvements, if any. 

We developed two versions of \cbert: one using the classes in Table~\ref{tab:care_definitions}, and a simpler one using  only the classes {\sc positive}, and {\sc negative}. The first four rows in Table~\ref{tab:care_definitions} are considered positive while the last three are negative, the results of which are featured in Table~\ref{tab:pos_neg_bert_results}. Naturally, the two-class model that blurs the differences between classes with the same valence has higher results.


\section{Modeling Analysis}

\begin{table}[H]
\centering
\small
\begin{tabular}{ccccc} 
\toprule
\textbf{} & Human & CARE & \cbert \\
\midrule
Human & 1.0 & 0.55 & 0.51\\
CARE & 0.89 & 1.0 & 0.72\\
\cbert & 0.72 & 0.62 & 1.0\\
\bottomrule
\end{tabular}
\caption{Percentage of agreement between annotation schemes. Each entry corresponds to the percentage of all labels the annotation scheme along the row agrees with the annotation scheme along the column.}
\label{tab:care_bert_stats}
\end{table}

\begin{table*}[htb]
\caption{Examples of posts labeled according to human annotators, CARE, and \cbert. The first three show examples where all three labeling schemes agree, the second three demonstrates examples where external knowledge may be needed, and the last three shows examples where the trajectory of the discussion may be more unpredictable. \textbf{Note: \cbert\  does not get access to the comments.}}
\small
\centering
\begin{tabular}{p{0.25\textwidth} | p{0.35\textwidth} | p{0.08\textwidth}| p{0.08\textwidth}| p{0.15\textwidth} }
    \toprule
      \hfil \textbf{Post} & \hfil \textbf{Comments} & \hfil \textbf{Human} & \hfil \textbf{CARE} & \hfil \textbf{\cbert} \\
      \toprule 
    Anxiety: I just want to say that I'm trying...I may not be successful, but I'm trying. & Dude, very proud of you my friend. Don't give up.; Good for you. I'm proud of you for trying. Keep at it.; Happy for you. & approving & approving & approving \\
    & & & & \\
    AskReddit: What's something you've been wanting to get off your chest but are too scared to? & I'm scared to end up alone and unloved.; I'm so scared to graduate college.; I just got engaged, but I'm not actually happy about it. & saddened; scared & saddened; scared & saddened; scared \\
    & & & & \\
    AskReddit: What movie really emotionally impacted you? & A Walk to Remember. So sad.; It's a Wonderful Life. Makes me so teary-eyed.; Dead Ringer. Made me so depressed. & saddened; approving & saddened; approving & saddened; approving\\
     \midrule
     Hockey: The Vancouver Canucks have landed a spot in the playoffs! & This is excellent news!; Holy shit, this is exciting!; Hell yeah, fuck the kings! & angered; excited; approving & excited & excited \\
     & & & & \\
    Panthers: Divisional Playoffs - Panthers vs. 49ers - Discussion Thread Let's do this! & I'M SO MAD; I'm freakin' scared, man.; Screw the whiners! They're going to regret the day they stepped on our turf! & angered; approving & angered & angered; excited \\
    & & & & \\
    InfertilityBabies: Going to be a line jumper! The doctor says my BP isn't stellar so I am in L and amp;D until Monday morning (37 weeks) induction!  & Good luck! So exciting!; Congrats! You're about to be a mom! I'm very excited for you!!!; Super exciting! & excited & excited & approving\\
    \midrule
    AskReddit: What is your favorite TV series ever? & Arrow. It's amazing!; Walking Dead. So excited for the new season!; Teen Titans. It's the best show ever. & approving; excited & approving & approving; amused \\
    & & & & \\
     Hearthstone: Who is this LIRIK guy, and why does he have 50K subscribers?
    & This is hilarious; What an idiot. Do more research before posting; He's an adorable guy. & amused; angered; adoring & amused & excited\\
    & & & & \\
    AskReddit: Imagine that the last thing you ate has been made illegal. What would that be? & Pizza, and now I'm super sad.; Frozen lasagna. Good riddance.; French onion dip. I love that stuff. & approving; saddened & saddened & approving  \\
      \bottomrule

\end{tabular}
 \newline
\label{tab:care_bert_examples}
\end{table*}

\end{document}